\documentclass{article}
\usepackage[preprint]{neurips_2025}
\usepackage{graphicx} %
\usepackage[utf8]{inputenc}
\usepackage{amsmath}
\usepackage{amssymb}
\usepackage{bm}
\usepackage{amsfonts}
\usepackage{stmaryrd}
\usepackage{mathtools}
\usepackage{hyperref}
\usepackage{cleveref}
\usepackage{subcaption}
\usepackage{color}
\usepackage{placeins}
\usepackage{booktabs}
\usepackage{svg}
\usepackage[table]{xcolor}
\usepackage{caption} 
\usepackage{pifont}
\usepackage{amsmath,amsfonts,bm}

\renewcommand{\vec}[1]{\mathbf{#1}}
\newcommand{\mat}[1]{\mathbf{#1}}
\newcommand{\Reals}{\mathbb{R}}

\newcommand{\FFN}{\text{FFN}}
\newcommand{\SAE}{\text{SAE}}

\newcommand{\W}{\mat{W}}

\newcommand{\Win}{\W_\text{1}}
\newcommand{\Wout}{\W_\text{2}}

\newcommand{\x}{\vec{x}}
\newcommand{\y}{\vec{y}}

\newcommand{\sx}{\vec{s}_\text{x}}
\newcommand{\sy}{\vec{s}_\text{y}}

\newcommand{\barsx}{\bar{\vec{s}}_\text{x}}

\newcommand{\ency}{e_\text{y}}
\newcommand{\decx}{d_\text{x}}

\newcommand{\Wdecx}{\mathbf{W}_\text{x}^\text{dec}}
\newcommand{\bdecx}{\mathbf{b}_\text{x}^\text{dec}}

\newcommand{\Wency}{\mathbf{W}_\text{y}^\text{enc}}
\newcommand{\bency}{\mathbf{b}_\text{y}^\text{enc}}

\newcommand{\Wdecxmj}{W_{\text{x},mj}^\text{dec}}
\newcommand{\Wencyik}{W_{\text{y},ik}^\text{enc}}

\newcommand{\Wencyactive}{\mathbf{W}_\text{y}^\text{enc(active)}}
\newcommand{\Wdecxactive}{\mathbf{W}_\text{x}^\text{dec(active)}}

\newcommand{\dimx}{m_\text{x}}

\newcommand{\dimy}{m_\text{y}}

\newcommand{\topk}{\tau_k}

\def\eqref#1{equation~\ref{#1}}

\def\1{\bm{1}}

\DeclareMathAlphabet{\mathsfit}{\encodingdefault}{\sfdefault}{m}{sl}
\SetMathAlphabet{\mathsfit}{bold}{\encodingdefault}{\sfdefault}{bx}{n}

\makeatletter
\newcommand{\IfPreprint}[2]{\if@preprint #1\else #2\fi}
\newcommand{\OnlyPreprint}[1]{\if@preprint #1\fi}
\newcommand{\OnlyAnon}[1]{\if@preprint\else #1\fi}
\makeatother

\title{SCALAR: Benchmarking SAE Interaction Sparsity in Toy LLMs}

\author{%
  Sean P. Fillingham$^*$ \\
  \texttt{spfillingham@gmail.com} \\
  \And
  Andrew Gordon$^*$ \\
  \texttt{algo2217@gmail.com} \\
  \And
  Peter Lai$^*$ \\
  \texttt{peterjlai@gmail.com} \\
  \And
  Xavier Poncini$^*$ \\
  \texttt{xponcini@gmail.com} \\
  \And
  David Quarel$^*$ \\
  Department of Computing \\ 
  Australian National University \\
  \texttt{david.quarel@anu.edu.au} \\
  \And
  Stefan Heimersheim\footnotemark[2] \\
  FAR.AI \\
  \texttt{stefan@far.ai} \\
}
\date{May 2025}

\begin{document}

\maketitle

\renewcommand{\thefootnote}{\fnsymbol{footnote}} %
\footnotetext[1]{These authors contributed equally to this work and are listed in alphabetical order.} %
\footnotetext[2]{Mentored this project while at Apollo Research.} %

\renewcommand{\thefootnote}{\arabic{footnote}}
\setcounter{footnote}{0}

\begin{abstract}
    Mechanistic interpretability aims to decompose neural networks into interpretable features and map their connecting circuits. The standard approach trains sparse autoencoders (SAEs) on each layer's activations. However, SAEs trained in isolation don't encourage sparse cross-layer connections, inflating extracted circuits where upstream features needlessly affect multiple downstream features. Current evaluations focus on individual SAE performance, leaving interaction sparsity unexamined. We introduce SCALAR (\textbf{S}parse \textbf{C}onnectivity \textbf{A}ssessment of \textbf{L}atent \textbf{A}ctivation \textbf{R}elationships), a benchmark measuring interaction sparsity between SAE features. We also propose "Staircase SAEs", using weight-sharing to limit upstream feature duplication across downstream features. Using SCALAR, we compare TopK SAEs, Jacobian SAEs (JSAEs), and Staircase SAEs. Staircase SAEs improve relative sparsity over TopK SAEs by $59.67\% \pm 1.83\%$ (feedforward) and $63.15\% \pm 1.35\%$ (transformer blocks). JSAEs provide $8.54\% \pm 0.38\%$ improvement over TopK for feedforward layers but cannot train effectively across transformer blocks, unlike Staircase and TopK SAEs which work anywhere in the residual stream. We validate on a $216$K-parameter toy model and GPT-$2$ Small ($124$M), where Staircase SAEs maintain interaction sparsity improvements while preserving feature interpretability. Our work highlights the importance of interaction sparsity in SAEs through benchmarking and comparing promising architectures.
\end{abstract}

\section{Introduction}
Large language models (LLMs) have achieved impressive capabilities, but their inner workings remain opaque.
Mechanistic interpretability seeks to explain model behavior by decomposing activations into interpretable
components called features, often using Sparse Autoencoders 
(SAEs) \citep{bricken2023monosemanticity, cunningham2023sparse, templeton2024scaling} or similar sparse dictionary
learning methods (e.g. Crosscoders \citep{lindsey2024crosscoders}, CLTs \citep{dunefsky2024transcoders}). 
These features can be organized
into circuits (graphs of interacting features) that trace how information flows through the model. The
sparsity of such circuits is important because simpler, sparser graphs offer clearer, more interpretable
explanations of model behavior.

However, existing Sparse Autoencoders (SAEs), which are widely used for extracting interpretable features,
do not necessarily yield sparse inter-layer interactions. Most SAEs are trained independently per layer,
leading to inconsistencies: a feature that is represented in one layer may be omitted (feature completeness,
\citep{bricken2023monosemanticity}), split into fine-grained features (feature splitting, 
\citep{bricken2023monosemanticity}), or absorbed into correlated
features (feature absorption, \citep{chanin2024featsplit}) on downstream layers. 
This introduces spurious connections between
features that are artifacts of the SAE training process rather than real interactions in the model. As a result,
they distort the circuit structure and hinder a simpler interpretation.

Recent work has attempted to address such issues through introducing Jacobian SAEs \citep[JSAEs,][]{farnik2025jacobian}.
These SAEs explicitly minimize inter-layer interaction by penalizing the L1 norm of the Jacobian between two adjacent
SAE latent vectors. While effective, this approach only applies to pairs of SAEs separated by a single feedforward layer.

To enable sparsification across a wider variety of components, we introduce \textit{Staircase SAEs},
which use a simple architectural
modification to encourage having fewer inter-layer feature interactions. The core idea is to encourage consistent
feature representation across layers by explicitly sharing upstream SAE weights in downstream SAEs. Concretely,
at a given layer, we set the dictionary of an SAE to be the concatenation of all upstream SAE features plus a 
new set of features
for the current layer. This causes the SAE width to grow in a staircase-like fashion while keeping the number 
of trainable parameters the same as a standard SAE. We illustrate this architecture in \Cref{fig:sae_staircase}.
As a result, features that are represented once can persist across layers, 
reducing spurious interactions and simplifying circuit structure—particularly for pass-through features flowing 
through the residual stream.

Jacobian SAEs and Staircase SAEs prompt the need for a quantitative evaluation of their
effectiveness, which we provide through a new benchmark: SCALAR
(Sparse Connectivity Assessment of Latent Activation Relationships). SCALAR quantifies
how performance degrades as inter-layer feature connections are ablated, similar to
established circuit discovery methods \citep[e.g.][]{conmy2023acdc}. We
sort connections between a pair of SAEs by their importance (measured using integrated
gradient attributions \cite{wang2024gradientbasedfeatureattribution}) and progressively remove them, 
measuring the KL divergence between
the original model output and the ablated output. We then compute the area under the degradation curve, which quantifies how compact the
interaction graph can be while preserving downstream performance.

To enable fair comparisons across different SAE configurations, we report both
absolute and relative interaction sparsity. Absolute sparsity measures the raw
area, while relative sparsity normalises area by the total number of connections between SAE pairs. Each metric on its own can be misleading: relative
sparsity can be inflated by adding inactive features, and absolute sparsity can favor
small but densely connected dictionaries.

\paragraph{Contributions}
Our main contributions in this work are as follows:
\begin{itemize}
    \item \textbf{Staircase SAEs}: We introduce Staircase SAEs, which use an architectural modification to
    improve the representation of pass-through circuits. Staircase SAEs simplify pass-through connections
    by explicitly sharing upstream features across all downstream SAEs. We find that, at any given
    layer, the active features consist of around 25 - 45\% new features and 55 - 75\% features
    from previous layers. This is compatible with the results of 
    \citet{balcells2024evolutionsaefeatureslayers} which
    studied pass-through circuits in standard SAEs.
    \item \textbf{SCALAR Benchmark}: We introduce SCALAR (Sparse Connectivity Assessment of Latent
    Activation Relationships), a benchmark for quantifying interaction sparsity between pairs of
    SAEs. SCALAR measures how model performance degrades as inter-layer connections are
    progressively ablated in order of importance, offering a principled measure of circuit simplicity.
    \item \textbf{Empirical Validation}: We use SCALAR to evaluate multiple SAE designs and find
    a tradeoff between standard SAEs, which may perform well initially but degrade
    under ablation, and Jacobian or Staircase SAEs, which retain performance even under pruning.
\end{itemize}

\section{Related Work}

\paragraph{SAE Architectures and Cross-Layer Interactions} Our work builds on research using SAEs to interpret neural networks \citep{cunningham2023sparse, bricken2023monosemanticity, gao2024scaling}. Current SAE methods face issues like feature splitting, absorption, and incompleteness that vary across layers, leading to inconsistencies in circuit interpretation. To address cross-layer interactions, \citet{farnik2025jacobian} introduced Jacobian SAEs (JSAEs) that minimize interaction sparsity via L1 Jacobian penalties, though limited to feedforward layers. \citet{yun2023transformer} and \citet{lindsey2024crosscoders} proposed shared SAEs and crosscoders, while \citet{dunefsky2024transcoders} introduced transcoders for cross-layer reconstruction.

\paragraph{Focus on Single-Layer Reconstruction} Although shared architectures and crosscoders yield sparse feature dictionaries across layers, they obscure individual layer representations. Since our goal is understanding computation flow within networks, we focus on approaches that reconstruct activations at single layers while improving cross-layer sparsity through architectural innovations rather than abandoning per-layer interpretability.

\section{Methodology}
We use a toy language model to benchmark the interaction sparsity of different SAE architectures: 
A 216K-parameter transformer \citep{vaswani2023attention} following the GPT-2 architecture \citep{Radford:19}. 
The model has 4 layers, a residual stream width of 64, and uses an ASCII tokenizer (vocabulary size of 128) 
similar to \citet{lai2025sae}.
We use a DynamicTanh layer \citep{zhu2025transformers} instead of LayerNorm to make computing 
the Jacobian for JSAEs tractable (see \Cref{app:jacobian_mlp_block}).
We train the model on the tiny-Shakespeare dataset \citep{karpathy2015unreasonable,karpathy2015shakespeare}
to perform standard next-token prediction.

SAEs are trained to reconstruct model activations that are mapped to and from
a sparse high dimensional space \citep{cunningham2023sparse,bricken2023monosemanticity}.
We train SAEs on activations between
transformer blocks, and on the input/output of each feedforward 
layer. In all cases, the activations are vectors of the same size as the 
residual stream. We train three types of SAEs: Conventional TopK SAEs
\cite{gao2024scaling}, Jacobian SAEs \citep[JSAEs][]{farnik2025jacobian},
and our new Staircase SAEs.

\subsection{TopK SAEs}

\label{sec:topk}
TopK SAEs use a TopK activation function which selects the $K$ largest magnitude
entries in the latent vector and sets all other entries to zero.
\begin{align}
	\mathbf{z} = \text{TopK}(\text{ReLU}(\mathbf{W}_{enc}(\mathbf{h} - \mathbf{b}_{dec}) + \mathbf{b}_{enc})),\quad
	\mathbf{z'} = \mathbf{W}_{dec} \mathbf{z} + \mathbf{b}_{dec}
\end{align}

Sparsity is enforced directly by the TopK operation, ensuring by 
definition\footnote{It may be less than $K$, due the ReLU.} 
that $||\mathbf{z}||_0 \leq K$. The SAEs are trained on the reconstruction loss
\begin{align}
	L_{\rm TopK\ SAE} = ||\mathbf{h} - \mathbf{h}'||_2^2.
\end{align}
We train TopK SAEs independently for each position in the model.

\subsection{Jacobian SAEs}
\label{subsec:jsae}

JSAEs \citep{farnik2025jacobian} sparsify the interactions between latents, in addition
to the latents themselves. This is done by imposing an L1 penalty on the Jacobian of
downstream latents with respect to upstream latents.
\begin{gather}
    L_{\rm Jac} = \Vert J \Vert_1,\quad J_{ij} = \frac{\partial z^{\rm downstream}_i(\mathbf{z^{\rm upstream}})}{\partial z^{\rm upstream}_j}.
\end{gather}
Due to the sparsity of the latents, the Jacobian itself is sparse,
and for a pair of TopK SAEs trained across a feedforward layer, there 
are only at most $K^2$ non-zero entries. These entries 
can be computed cheaply in closed form, see \Cref{app:jacobian_mlp_block}. 
JSAEs are trained with a mixture of both losses: 
\begin{align}
L_{\rm JSAE} = L_{\rm TopK\ SAE} + \lambda L_{\rm Jac}
\end{align}

Applying JSAEs to standard Transformer feedforward blocks that use LayerNorm (LN) \citep{ba2016layernorm} presents 
a challenge due to LN's inherent dependencies between elements of the input vector. 
\begin{align}
	{\rm FFBlock}(x) = x + {\rm FF}({\rm LN}(x))
\end{align}
These dependencies complicate the modification of the efficient closed-form of $J$ for feedforward blocks,
resulting in a large intermediate term when computing the Jacobian (\Cref{app:jacobian_derivation}). 
In contrast, DynamicTanh (DyT) \citep{zhu2025transformers} is an element-wise operation, 
that can be used as a drop-in replacement for LN. We fine-tune our 
model to substitute LN layers with DyT (\Cref{app:dyt}), modify the derivation of $J$ in \citet{farnik2025jacobian} 
to incorporate both DyT and the skip connection, and utilize this Jacobian to train JSAEs over MLP blocks.

\subsection{Staircase SAEs}
\label{subsec:staircase_sae}

We introduce the \textbf{Staircase SAE}, 
an architecture aimed at promoting feature reuse from earlier layers to aid in activation reconstruction. 
This approach seeks to facilitate circuit sparsity through structural design rather than encouraging it
via an explicit training objective (as for JSAEs).

\begin{figure}[htbp]
	\centering
	\begin{minipage}[t]{0.48\linewidth}
		\centering
		\includegraphics[width=0.8\linewidth]{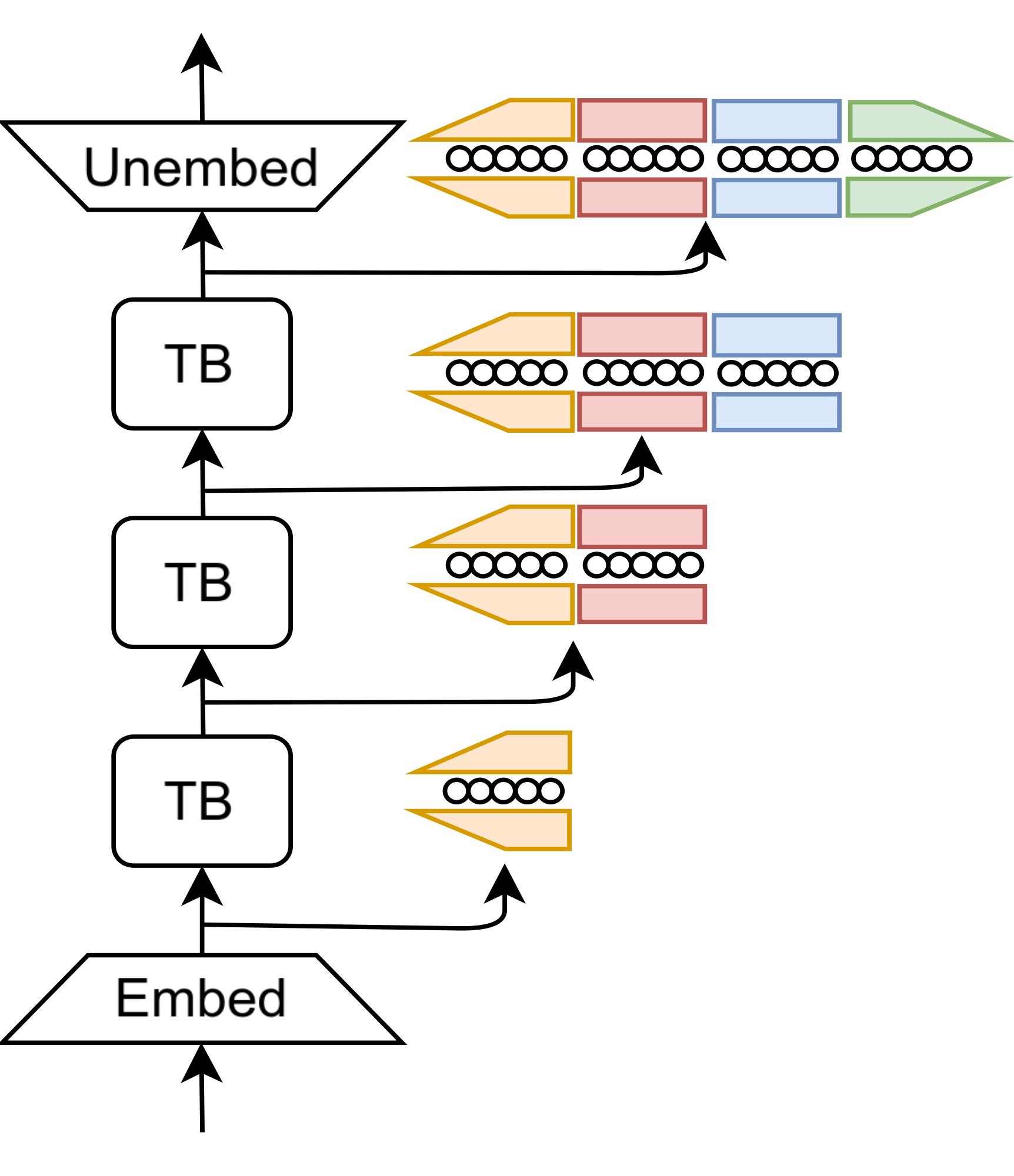}
		\caption{The staircase SAE architecture for a transformer with $L=3$ layers. Each layer $i$ uses a slice of the
		shared encoder $\mathbf{W}_{enc}$ and decoder $\mathbf{W}_{dec}$ weights. SAE chunks of identical
		colour indicate weights shared within the slices $\mathbf{W}^i_{enc}$ and $\mathbf{W}^i_{dec}$.}
		\label{fig:sae_staircase}
	\end{minipage}
	\hfill
	\begin{minipage}[t]{0.48\linewidth}
		\centering
		\includegraphics[width=\linewidth]{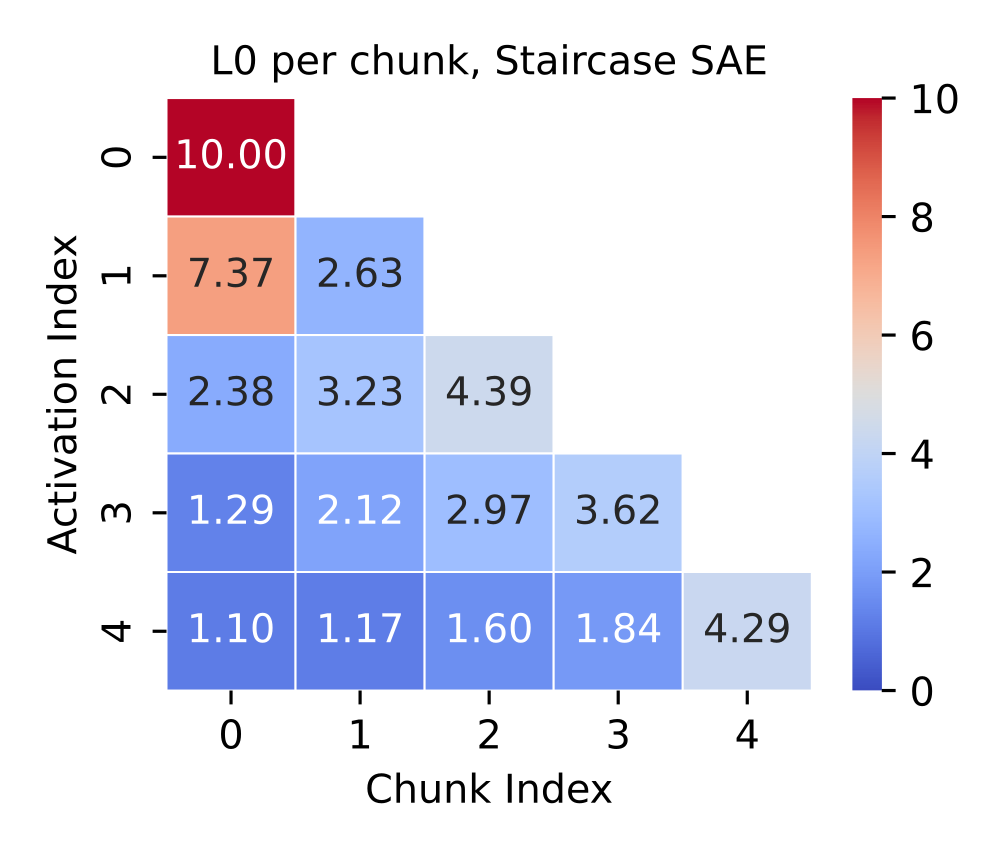}
		\caption{By measuring the number of active latents per chunk,
		we can see feature reuse from previous layers, as each SAEs allocates
		some ``sparsity budget'' to features from previous layers.}
		\label{fig:l0_staircase}
	\end{minipage}
\end{figure}

\label{sec:staircase_sae}
Staircase SAEs form a collection of related SAEs that use shared weights. All SAEs use the same encoder and decoder weight matrices, $\mathbf{W}_{enc}$ and $\mathbf{W}_{dec}$. Each SAE layer $i$ (for $1 \leq i \leq L+1$) accesses a progressively larger slice of these shared weights. Assuming a base feature chunk size of $n$, the feature dimension for an SAE on layer $i$ is $n \times i$. The encoder and decoder weights for layer $i$ are:
\begin{align}
	\mathbf{W}^i_{enc}  := \mathbf{W}_{enc}[:\;ni, :] \in \mathbb{R}^{d_{\text{model}} \times ni}, \quad 
	\mathbf{W}^i_{dec}  := \mathbf{W}_{dec}[:, :ni] \in \mathbb{R}^{ni \times d_{\text{model}}}
\end{align}
where $\mathbf{W}_{enc} \in \mathbb{R}^{d_{\text{model}} \times N}$ 
and $\mathbf{W}_{dec} \in \mathbb{R}^{N \times d_{\text{model}}}$ are the full 
shared matrices (see \Cref{fig:sae_staircase}). The notation $\mathbf{M}[:k, :]$ selects the 
first $k$ rows of $\mathbf{M}$; $\mathbf{M}[:, :k]$ selects the first $k$ columns. Each successive layer 
thus gains access to all earlier features, plus an additional chunk of $n$ new features.

Crucially, the biases $\mathbf{b}_{enc}^i \in \mathbb{R}^{ni}$ 
and $\mathbf{b}_{dec}^i \in \mathbb{R}^{d_{\text{model}}}$ are \textit{independent} for each layer $i$. 
Independent decoder biases $\mathbf{b}_{dec}^i$ allow each SAE to center activations appropriately for its layer. 
Independent encoder biases $\mathbf{b}_{enc}^i$ are essential for feature selection and reuse: 
if features from earlier chunks ($j < i$) are unhelpful for reconstructing layer $i$, the optimizer can 
suppress them by setting the corresponding bias entries to large negative values (assuming ReLU-like activation). 
With shared biases, disabling a feature for one layer could inadvertently degrade performance on another where that feature is important.

A Staircase SAE is at least as expressive as a set of independent SAEs of width $n$, while using nearly the 
same number of parameters.\footnote{The only additional parameters are the per-layer bias terms $\mathbf{b}_{enc}^i$ 
and $\mathbf{b}_{dec}^i$, which increased the total by only $\approx 1.5\%$ in our experiments.} This equivalence 
can be achieved by setting $\mathbf{b}_{enc}^i[0:n(i{-}1)] := -\infty$, ensuring that only the $i$-th chunk of 
features $\mathbf{z}[n(i{-}1):ni]$ is active and effectively disabling features from earlier chunks.

To verify that Staircase SAEs behave differently from independent SAEs and promote feature reuse, we measured 
$L_0$ sparsity across chunks when using the Staircase architecture with the TopK SAE variant. While each layer 
tends to preferentially use features from its own chunk ($i$-th chunk for layer $i$), each layer also reuses 
features from earlier chunks ($0$ to $i{-}1$) to improve reconstruction (\Cref{fig:l0_staircase}).

We also evaluated an alternative SAE architecture in which each layer $i$ could only optimize the weights 
of its own chunk, relying on earlier layers ($j < i$) to make features in previous chunks useful. 
This design underperformed even standard SAEs (\Cref{app:staircase_detach}), underscoring the importance of 
allowing each layer to directly optimize all accessible weights. 

The Staircase architecture is compatible with any SAE variant and can be used with Standard, TopK, or other 
formulations. A comparison with other variants in \Cref{app:staircase_variants} shows that the Staircase 
architecture enhances the TopK SAE variant using a negligible increase in parameter count.

\subsection{Interaction sparsity benchmark: SCALAR}
\label{subsec:SCALAR}

In this section, we describe how SCALAR scores are computed. This methodology is architecture-agnostic, 
so we proceed abstractly. Suppose we have a language model and a pair of SAEs, 
denoted $\text{SAE}_0$ (upstream) and $\text{SAE}_1$ (downstream), where the upstream SAE encodes
and decodes at an earlier layer in the model than the downstream one. We refer to the original, 
unmodified model as the \textit{full model}, and to the version where activations at the SAE positions 
are replaced with their reconstructions as the \textit{full circuit}. For clarity, we use \textit{latent} 
to refer to an index in an SAE latent vector, and \textit{latent magnitude} for its corresponding value. A SCALAR score is computed using the following steps:

\begin{enumerate}
    \item \textbf{Scoring connections}: For each latent $i$ in $\text{SAE}_0$ and each latent 
	$j$ in $\text{SAE}_1$, measure how strongly $i$ affects $j$. This produces a list of index 
	pairs $(i, j)$, referred to as \textit{edges}, ordered by interaction strength.
    
    \item \textbf{Cutting connections}: Choose a sequence of values representing edge counts, called 
	the \textit{edge number sequence}. For each $n$ in the sequence, retain only the top $n$ edges 
	(by interaction strength) and ablate the rest. The resulting model, with only a subset of connections 
	preserved, is called a \textit{subcircuit}.
    
    \item \textbf{Ablation curve}: For each $n$ and a batch of prompts, compute the Kullback–Leibler 
	(KL) divergence between the logits of the full model and those of the subcircuit. Average the KL 
	divergence across sequence positions and prompts to obtain a single value per $n$. 
	The resulting values define the \textit{ablation curve}, a piecewise linear function mapping edge 
	count to KL divergence.
    
    \item \textbf{Area under the curve}: The \textit{absolute SCALAR score} is the area under the 
	ablation curve. The \textit{relative SCALAR score} is this area normalized by the total number 
	of edges between $\text{SAE}_0$ and $\text{SAE}_1$.
\end{enumerate}

By measuring the KL divergence between the logits of the full model and those of the subcircuit, 
the SCALAR score captures both the computational sparsity and reconstruction quality of an SAE pair. 
With a simple modification, a SCALAR score can also be used to assess computational sparsity alone, see Appendix \ref{app:PCSM} for discussion. 

\paragraph{Intuitive Example} Consider the sparse feature circuits from \citet{marks2025sfc}, where multiple "plural nouns" features across layers interact densely (their Figure 11). Such circuits would receive high (poor) SCALAR scores due to many important cross-layer connections. A more interaction-sparse architecture would consolidate these into fewer shared features, yielding lower (better) SCALAR scores and simpler, more interpretable circuits.

\subsubsection{Scoring Connections}
\label{subsubsec:SC}
Before cutting connections, we first identify which links between latents in $\text{SAE}_0$ and 
$\text{SAE}_1$ are most important to computation. To do this, we express downstream latent 
magnitudes as a function of upstream ones: we decode the upstream latents, advance the resulting 
activations through the model to the position of the downstream SAE, and then re-encode them 
into downstream latent magnitudes.

We then sample upstream latent magnitudes from data and, for each sample, use integrated gradients 
\cite{sundararajan2017} to estimate the importance of each upstream latent to each downstream latent. 
Repeating this across samples, we compute the root mean square of the attributions to rank the 
importance of each connection between the two layers.

\begin{figure}[t!]
    \centering
    \begin{subfigure}[b]{0.34125\textwidth}
        \centering
        \includegraphics[width=\textwidth]{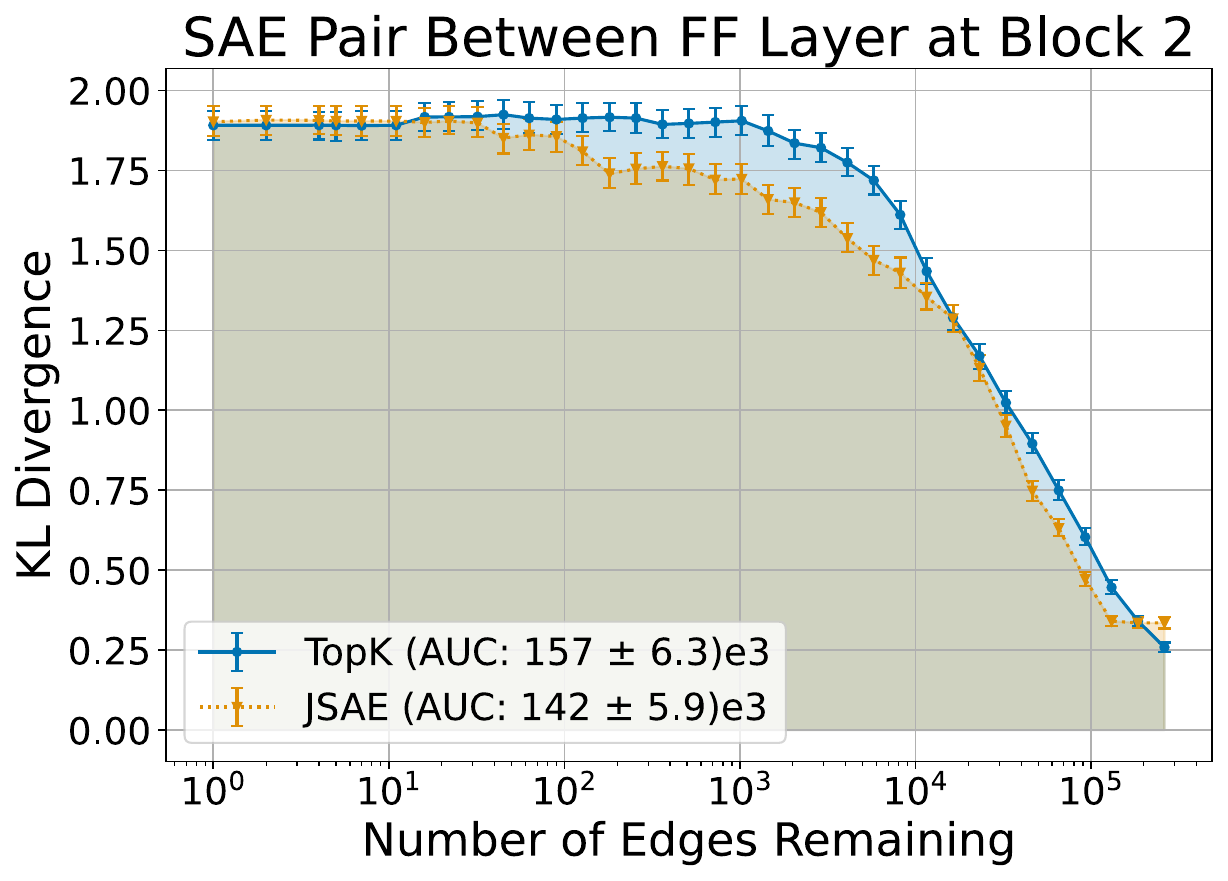}
    \end{subfigure}
    \hfill
    \begin{subfigure}[b]{0.31\textwidth}
        \centering
        \includegraphics[width=\textwidth]{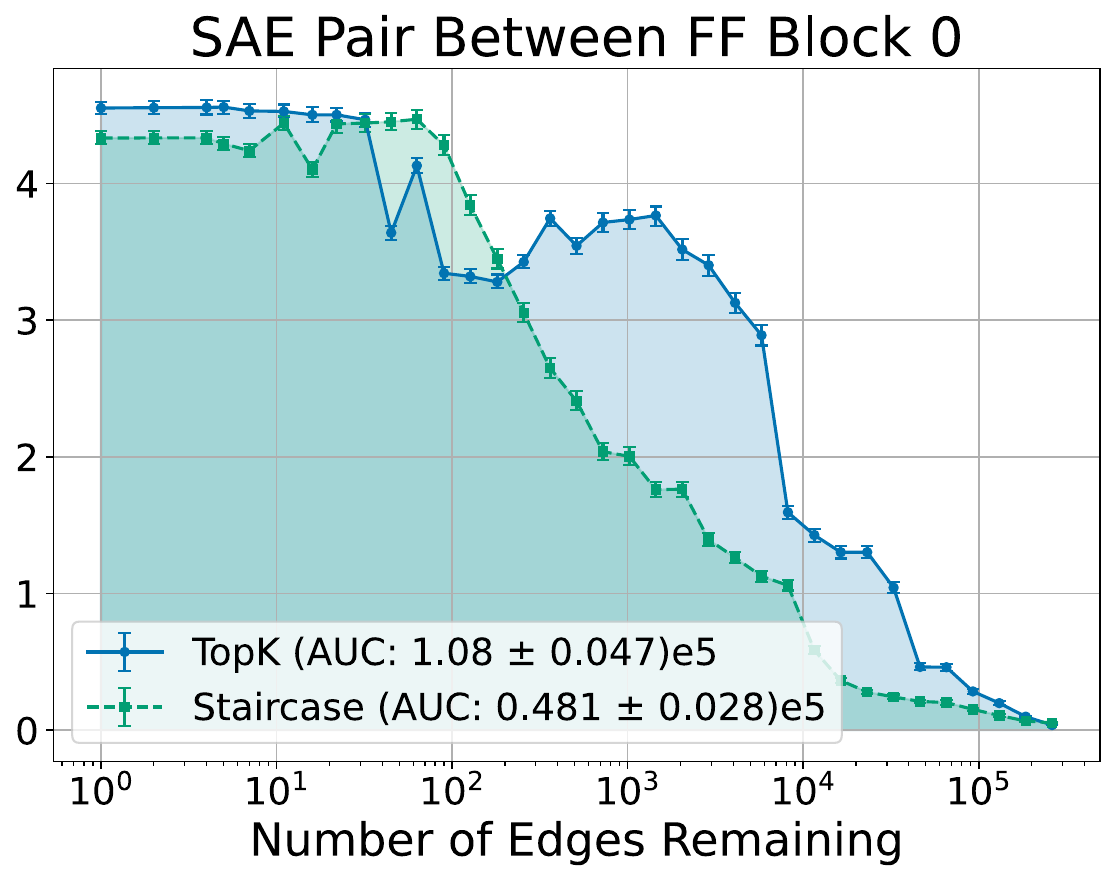}
    \end{subfigure}
    \hfill
    \begin{subfigure}[b]{0.31\textwidth}
        \centering
        \includegraphics[width=\textwidth]{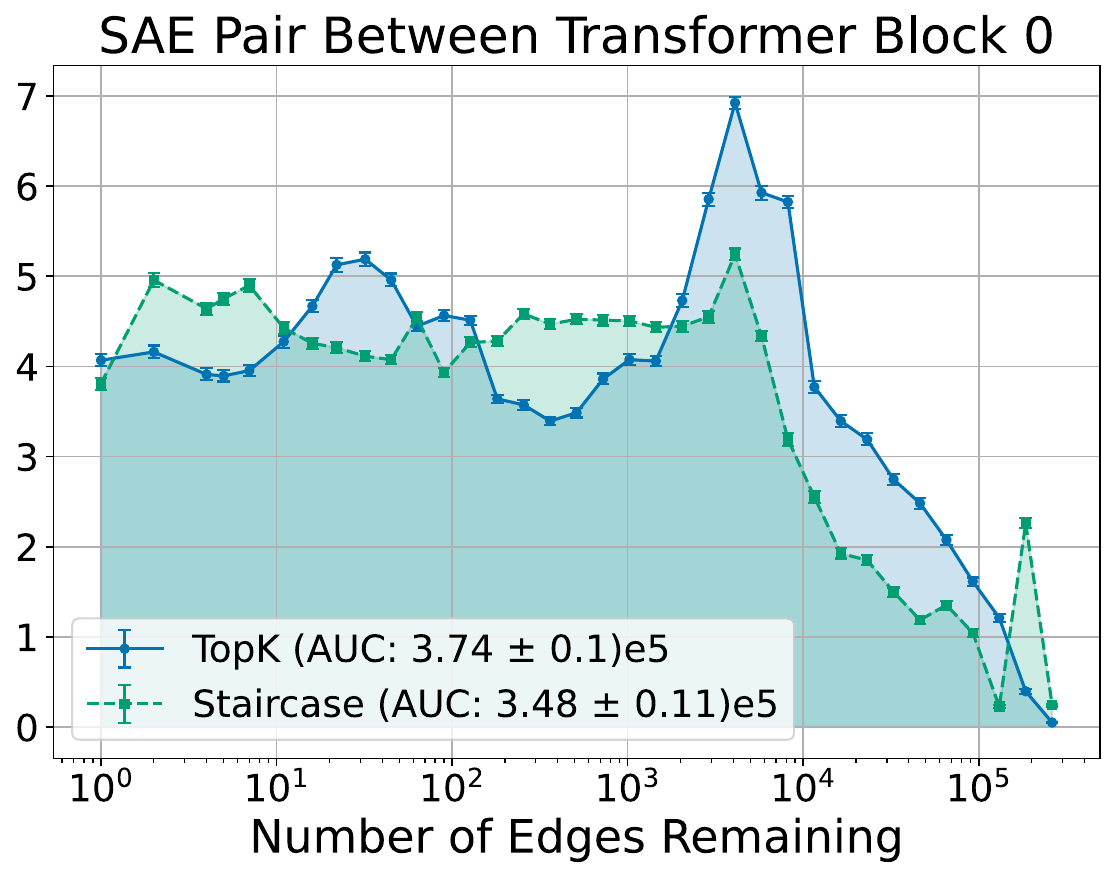}
    \end{subfigure}
    \caption{The ablation curves for all SAEs attached at the labeled compute block. 
	In these examples, the JSAE and Staircase SAEs clearly outperform the standard TopK SAEs.}
    \label{fig:three_figures_good}
\end{figure}

\subsubsection{Cutting Connections}

For clarity, we represent the full circuit as a \textit{bipartite graph} $S$, where vertices 
correspond to the latents of $\text{SAE}_0$ and $\text{SAE}_1$, which form disjoint sets. 
A subcircuit corresponds to a subgraph $T \subset S$. While a forward pass through the full circuit 
is straightforward (see Section~\ref{subsec:SCALAR}), we now describe how to compute a forward pass 
through the subcircuit $T$.

The subcircuit computes upstream latent magnitudes by applying the usual SAE encoding to model 
activations at the position of $\text{SAE}_0$. The downstream latents, however, are computed selectively. 
We first initialize an output tensor for the downstream latent magnitudes, setting all entries 
(across sequence positions) to zero. This tensor will be populated as follows: for each downstream 
latent $\ell$ included in $T$ (i.e., a node in $\text{SAE}_1$), we perform the following steps:

\begin{enumerate}
    \item \textbf{Identify upstream latents}: Let $U_{\ell}$ be the set of upstream 
	latents (i.e., $\text{SAE}_0$ nodes) connected to $\ell$.
    
    \item \textbf{Ablate irrelevant latents}: Zero out all upstream latent magnitudes not in 
	$U_{\ell}$ across all sequence positions.
    
    \item \textbf{Forward pass}: Decode the ablated upstream latents, run the resulting 
	activations through the model segment between the SAEs, and encode with $\text{SAE}_1$.
    
    \item \textbf{Update output}: Insert the resulting latent magnitude for $\ell$ at each 
	sequence position into the output tensor for the downstream latent magnitudes.
\end{enumerate}

\section{Results}
\label{sec:results}

\subsection{SAEs Across a Feedforward Layer}
We trained both TopK SAEs and JSAEs across each feedforward layer. 
The sparsity coefficient of the JSAE pairs are tuned for performance on our metric, see \Cref{app:jacobian_coeff_tuning}. 
Example ablation curves are shown in the leftmost panels of \Cref{fig:three_figures_good}; 
full curves for each layer are included in Appendix \ref{app:ablation_curves}.
In \Cref{fig:three_figures_good}, the JSAE curve lies below the TopK curve at compute block 2 (the third transformer block), 
indicating higher computational sparsity (i.e., fewer active edges are needed to reach the same KL divergence). 
Performance varies across layers, with JSAEs outperforming TopK SAEs in central blocks while TopK performs better in peripheral ones, 
as shown in \Cref{fig:SCALAR_scores_auc} where absolute SCALAR scores reflect this layer-dependent performance pattern. 
Cases where performance is mixed are shown in Appendix \ref{app:additional_ablation_curves}. 

The percentage change in SCALAR scores for each layer is reported in \Cref{tab:scalar_feedforwardlayer}, 
showing that JSAEs outperform TopK SAEs in central blocks, while TopK performs better in peripheral ones. 
We highlight the outlier performance of the JSAE pair at Layer 0 on percentage reduction, this is likely 
due to poor tuning of the Jacobian coefficient, see \Cref{app:jacobian_coeff_tuning}. 
Despite this, the value of the SCALAR score is insignificant compared to other layers, 
see \Cref{fig:SCALAR_scores_auc}. 
Overall, summing SCALAR scores across all layers (absolute and relative are equivalent in this case), 
JSAEs provides a $8.54\% \pm 0.38\%$ improvement over TopK.

\begin{table}[h]
    \centering
    \resizebox{\textwidth}{!}{\begin{tabular}{lccccc}
        \toprule
        & \textbf{Layer 0} & \textbf{Layer 1} & \textbf{Layer 2} & \textbf{Layer 3} & \textbf{Aggregate}\\
        \midrule
        \textbf{Absolute reduction (\%)}  & $-602.15 \pm 335.68$ & $24.50 \pm 1.57$ & $9.55 \pm 0.52$ & $-14.07 \pm 0.96$ & $8.54 \pm 0.38$\\
        \bottomrule
    \end{tabular}}
    \vspace{0.5em}
	\caption{Percentage reduction in SCALAR scores for JSAEs compared to TopK SAEs across feedforward layers. Positive values indicate improved sparsity (lower SCALAR score) for JSAEs.}
        \label{tab:scalar_feedforwardlayer}
\end{table}

\subsection{SAEs Across a Feedforward Block}
Across the entire feedforward block
we trained pairs of standard TopK SAEs and Staircase SAEs. 
The central panels of \Cref{fig:three_figures_good} show representative ablation curves where 
the Staircase SAE clearly outperforms TopK around compute block 0. However, performance varies across different compute blocks, 
with some showing mixed results (see Appendix \ref{app:additional_ablation_curves} for complete results).
These trends are reflected in \Cref{fig:SCALAR_scores_auc}, with Staircase achieving higher sparsity at block 0. 
While the Staircase architecture permits more potential connections, leading to higher 
absolute SCALAR scores, the relative SCALAR score still indicates improved sparsity. 

Layer-wise changes in SCALAR scores are shown in \Cref{tab:scalar_feedforwardblock}. Staircase SAEs yield lower absolute SCALAR scores at layer 0 but higher scores at layers 1-3. However, relative SCALAR scores are consistently lower across all layers. Summing over all layers, Staircase SAEs provide a $19.24\% \pm 0.71\%$ improvement in absolute SCALAR score and a $59.67\% \pm 1.83\%$ improvement in relative SCALAR score compared to TopK. This cumulative improvement, particularly with regards to the relative SCALAR score, suggests that Staircase SAEs achieve greater sparsity across the feedforward block.

\begin{table}[h]
    \centering
    \resizebox{\textwidth}{!}{\begin{tabular}{lccccc}
        \toprule
        & \textbf{Block 0} & \textbf{Block 1} & \textbf{Block 2} & \textbf{Block 3} & \textbf{Aggregate} \\
        \midrule
        \textbf{Absolute reduction (\%)} & $55.46 \pm 2.94$ & $-1.53 \pm 0.10$ & $-7.41 \pm 0.45$ & $-10.33 \pm 0.79$ & $19.24 \pm 0.71$\\
        \textbf{Relative reduction (\%)} & $77.78 \pm 3.74$ & $49.24 \pm 2.52$ & $46.50 \pm 2.23$ & $45.11 \pm 2.74$ & $59.67 \pm 1.83$\\
        \bottomrule
    \end{tabular}}
    \vspace{0.5em}
	\caption{Percentage reduction in SCALAR scores for Staircase SAEs compared to TopK SAEs across feedforward blocks. Positive values indicate improved sparsity (lower SCALAR score) for Staircase SAEs.}
        \label{tab:scalar_feedforwardblock}
\end{table}

\subsection{SAEs Across a Transformer Block}
Across each transformer block,
we trained both standard TopK SAEs and Staircase SAEs. The rightmost panels of \Cref{fig:three_figures_good} show representative ablation curves where Staircase SAEs perform well. Staircase SAEs outperform TopK SAEs at some compute blocks with consistently lower ablation curves, while at other blocks the performance is mixed with curves crossing at different thresholds (see Appendix \ref{app:additional_ablation_curves}). \Cref{fig:SCALAR_scores_auc} reflects this variable pattern in the absolute SCALAR scores. However, when using the relative SCALAR score to account for the increased feature capacity of Staircase SAEs, the ranking consistently favors Staircase SAEs across blocks.

The percentage changes in the SCALAR scores are shown in \Cref{tab:scalar_transformer_block}. Staircase SAEs achieve lower absolute SCALAR scores in layer 0 and higher scores in layers 1-3, but consistently lower relative SCALAR scores across all layers. Aggregated across layers, Staircase SAEs yield a $-29.46\% \pm 0.91\%$ improvement in absolute SCALAR score and a $63.15\% \pm 1.35\%$ improvement in relative SCALAR score compared to TopK.
In summary, while the absolute SCALAR metric suggests TopK SAEs are sparser overall, the relative SCALAR score, which normalizes for model capacity, indicates that Staircase SAEs achieve greater effective sparsity across transformer blocks.

\begin{table}[h]
    \centering
    \setlength{\tabcolsep}{5pt}
    \resizebox{\textwidth}{!}{\begin{tabular}{lccccc}
        \toprule
        & \textbf{Layer 0} & \textbf{Layer 1} & \textbf{Layer 2} & \textbf{Layer 3} & \textbf{Aggregate}\\
        \midrule
        \textbf{Absolute reduction (\%)} & $6.95 \pm 0.28$ & $-64.50 \pm 3.69$ & $-107.42 \pm 7.16$ & $-125.53 \pm 12.84$ & $-29.46 \pm 0.91$\\
        \textbf{Relative reduction (\%)} & $53.48 \pm 1.63$ & $72.60 \pm 2.99$ & $82.79 \pm 2.55$ & $88.83 \pm 4.28$ & $63.15 \pm 1.35$\\
        \bottomrule
    \end{tabular}}
    \vspace{0.5em}
        \caption{Percentage reduction in SCALAR scores for Staircase SAEs compared to TopK SAEs across transformer blocks. Positive values indicate improved sparsity (lower SCALAR score) for Staircase SAEs.}
    \label{tab:scalar_transformer_block}
\end{table}

\begin{figure}[t!]
    \centering
    \includegraphics[width=\linewidth]{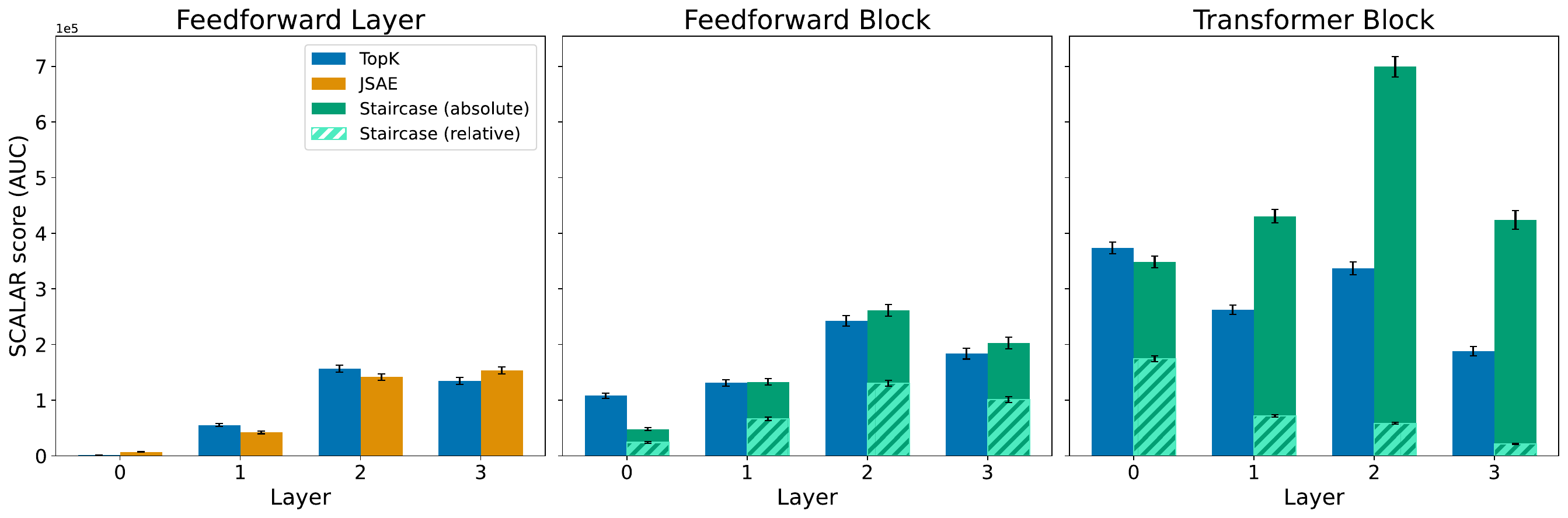}
    \caption{A comparison of SCALAR scores across SAE positions and variants. In this space a lower SCALAR score is suggestive of higher sparsity. So, for example, around the Transformer block at layer 1 the TopK SAE exhibits higher sparsity than the Staircase SAE using the absolute SCALAR score. However, at that same position, the Staircase SAE has higher sparsity when using the relative SCALAR score.}%
    \label{fig:SCALAR_scores_auc}
\end{figure}

\subsection{Validation on GPT-2 Small}
To test generalization to realistic model scales, we trained TopK and Staircase SAE pairs on GPT-2 Small (124M parameters). Using down-sampled integrated gradients and ablation studies on layers 1, 6, and 11, we find that Staircase SAEs provide a $38.7 \pm 1.2\%$ improvement over TopK in relative interaction sparsity. While this improvement is smaller than our toy model results ($59.67 \pm 1.83\%$), it demonstrates that our core findings generalize beyond the toy setting. Full details and results are provided in Appendix \ref{app:GPT-2}.

\subsection{Feature Interpretability Assessment}
To verify that interaction sparsity improvements don't compromise individual feature interpretability, we conducted a blinded study comparing 1000 features each from Staircase and TopK SAEs (GPT-2 Small, layer 7). Human evaluators preferred TopK features $348 \pm 15$ times, Staircase features $342 \pm 15$ times, and were indifferent $310 \pm 15$ times. This suggests Staircase SAEs 
are equally interpretable individually while achieving greater interaction sparsity, supporting our architectural approach to improving circuit interpretability.

\section{Discussion}
While our results demonstrate the promise of Staircase SAEs for improving circuit sparsity and usefulness of the SCALAR metric, several limitations and open questions remain. Below, we outline some areas for future work and clarify the scope of our current contributions.

\textbf{Model Scale Validation:} While our primary experiments use a 216K-parameter toy model, we validate our key findings on GPT-2 Small (124M parameters). The $38.7\%$ improvement in relative interaction sparsity, though smaller than our toy model results ($59.67\%$), provides evidence that our approach generalizes to realistic model architectures. Full technical details are provided in Appendix \ref{app:GPT-2}.

\textbf{Surrogate models vs. explaining activations:} In our work, we focus on SAEs instead of
more recent Crosscoders and Transcoders. One advantage of SAEs is that they show all information
present in a cross-section of the model – we know that features interactions between two SAE layers
must be mediated by the transformer block in between.

\textbf{Downstream uses of SAEs:} In this work, we focused on circuit interpretability conditioned on sparse dictionary learning and SAEs. Recent work however has called into question whether the apparent interpretability of SAEs is useful in downstream tasks \citep{kantamneni2025saenotuseful}. Our interpretability assessment suggests that architectural improvements to interaction sparsity need not compromise individual feature quality, addressing one potential concern about SAE utility.

\section{Conclusion}
Sparse autoencoders are widely used in mechanistic interpretability to disentangle superimposed model features into sparse, human-understandable latents; however, to fully trace how information flows through a model, we need not only interpretable features but also interpretable circuits with sparse and meaningful connections between features. Prior work has largely focused on per-layer sparsity and reconstruction loss, neglecting the sparsity of these inter-feature connections. Our work addresses this gap by characterizing interaction sparsity between adjacent SAEs trained on the residual stream of a toy LLM.

We introduce a general-purpose architecture, \textit{Staircase SAEs}, which improves interaction sparsity by allowing downstream layers to reuse upstream features. This design is compatible with all SAE variants and facilitates the emergence of pass-through circuits. To evaluate interaction sparsity, we propose the \textit{SCALAR score}, a metric that characterizes the importance of cross-layer connections. Using this score, we assess the circuit sparsity of standard TopK SAEs, Jacobian SAEs, and Staircase SAEs across various model layers. We find that TopK SAEs exhibit the lowest interaction sparsity, while both Jacobian and Staircase SAEs achieve higher sparsity under the relative SCALAR score.

One challenge with Staircase SAEs is that they inherently allow more potential cross-layer connections, which leads to higher absolute SCALAR scores. To account for this, we introduce the \textit{relative SCALAR score}, which normalizes for connection count and offers a fairer basis for comparison across architectures. Using this metric, we show that architectural choices can meaningfully shape circuit sparsity. We validate our approach on models ranging from 216K to 124M parameters, demonstrating that architectural choices can meaningfully improve interaction sparsity without sacrificing individual feature interpretability. We hope this work encourages continued exploration of SAE architectures that support sparser and more interpretable cross-layer interactions.

\section*{Acknowledgments}

We would like to thank Lucy Farnik for insightful discussions about Jacobian SAEs.
The authors would like to thank the Supervised Program for Alignment Research (SPAR) for operational support. 
PL conducted this research as part of the MARS program by the Cambridge AI Safety Hub (CAISH).
XP acknowledges support from the European Research Council (ERC) under the European Union's Horizon 2020 
research and innovation programme (101042460): ERC Starting grant ``Interplay of structures in conformal and universal 
random geometry'' (ISCoURaGe, PI Eveliina Peltola).

\FloatBarrier
\bibliography{references, references_david} %

\newpage

\newpage
\appendix
\section{Compute resources}
\label{sec:compute_resources}
The LLM and all SAEs were trained on a local machine with 4xA4000s, 64GB CPU RAM and 2TB of storage.
The LLM takes $\leq 5$ minutes to train, and each SAE takes $\approx 30$ mins to train. 
We estimate the total compute to be no more than 100 GPU hours to replicate
the results in the paper.

The gradient attributions were trained on Runpod, on a machine with 1xA40, 50 GB CPU RAM and 20GB of storage.
One set of attributions for a 4 layer model takes $\leq 20$ minutes to produce.

The SCALAR scores were computed on an Apple M4 machine with 24GB CPU RAM and 250GB of storage. Computing a SCALAR score for an SAE pair with edge count sequence (\ref{equ:edge_count_sequence}) across 50 prompts takes $\approx 100$ minutes to produce.

\section{Staircase Detached Variant}
\label{app:staircase_detach}

When training Staircase SAEs, the optimizer is allowed to optimize all weights
in each layer to minimize reconstruction loss. Looking at \Cref{fig:sae_staircase},
this means that the orange chunk sees activations $\mathbf{h}^0, \ldots, \mathbf{h}^3$ from between all layers,
the blue chunk sees $\mathbf{h}^1, \mathbf{h}^2, \mathbf{h}^3$, and so on, up to the green chunk which
only sees $\mathbf{h}^3$ during training.

One variant that was considered was to detach the gradients from the previous chunks used in subsequent layers,
so that the orange chunk would only be optimized to reconstruct $\mathbf{h}^1$. The optimizer for the subsequent layer
can reuse the feature magnitudes that the orange chunk computes if it is useful for reconstructing $\mathbf{h}^2$, but
is not allowed to directly update the orange chunk's weights.

\begin{figure}[htbp]
	\centering
	\includegraphics[width=\linewidth]{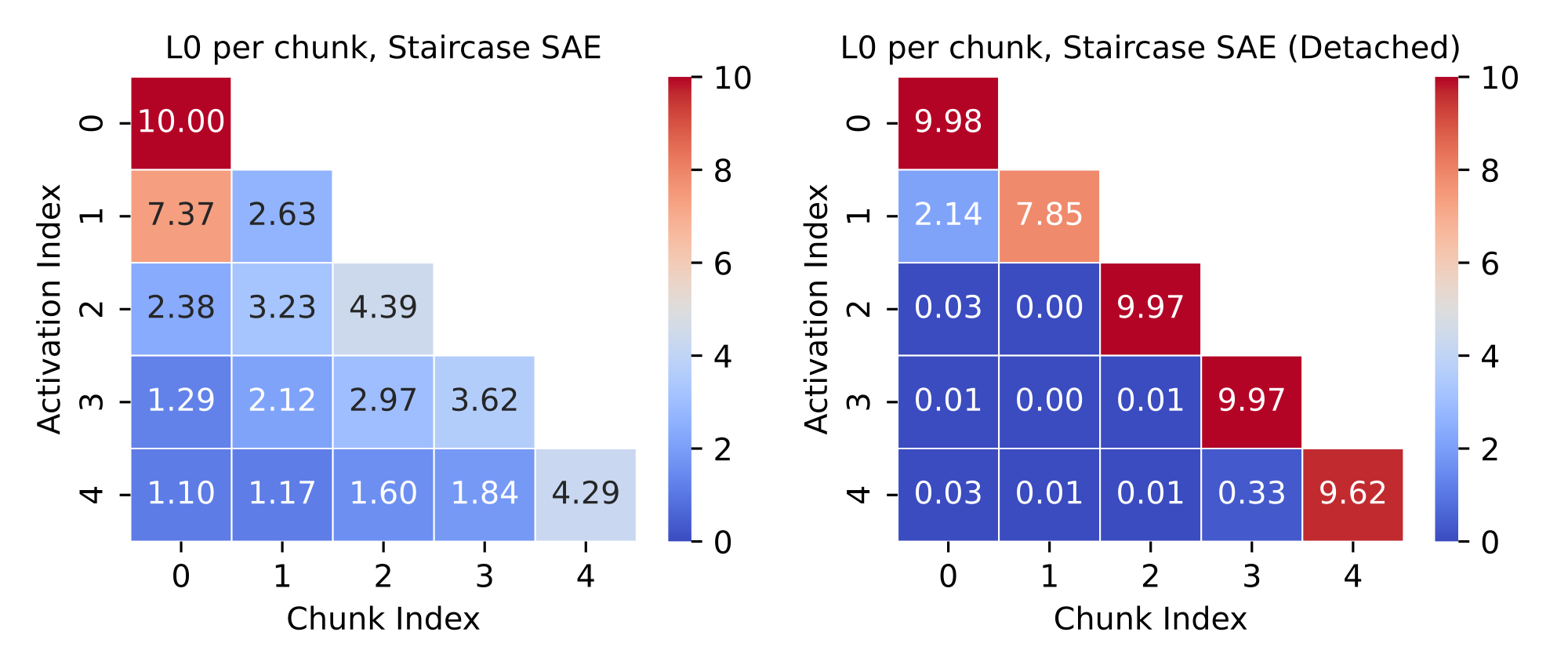}
	\caption{The L0 sparsity measured per chunk for a staircase SAE with $L=4$ layers,
	and 5 activations $\mathbf{h}^0, \ldots, \mathbf{h}^4$.
	The left figure was trained with all gradients attached, while the right figure was trained with
	gradients from previous chunks detached. Both models use Top-$10$ SAEs. What we find is that the standard
	staircase variant (left) spends some sparsity budget on features from previous chunks,
	whereas the detached gradient variant (right) degenerates back to a standard SAE, rarely using
	features from previous chunks.}
	\label{fig:l0_per_chunk}
\end{figure}

\begin{figure}[htbp]
	\centering
	\includegraphics[width=0.5\linewidth]{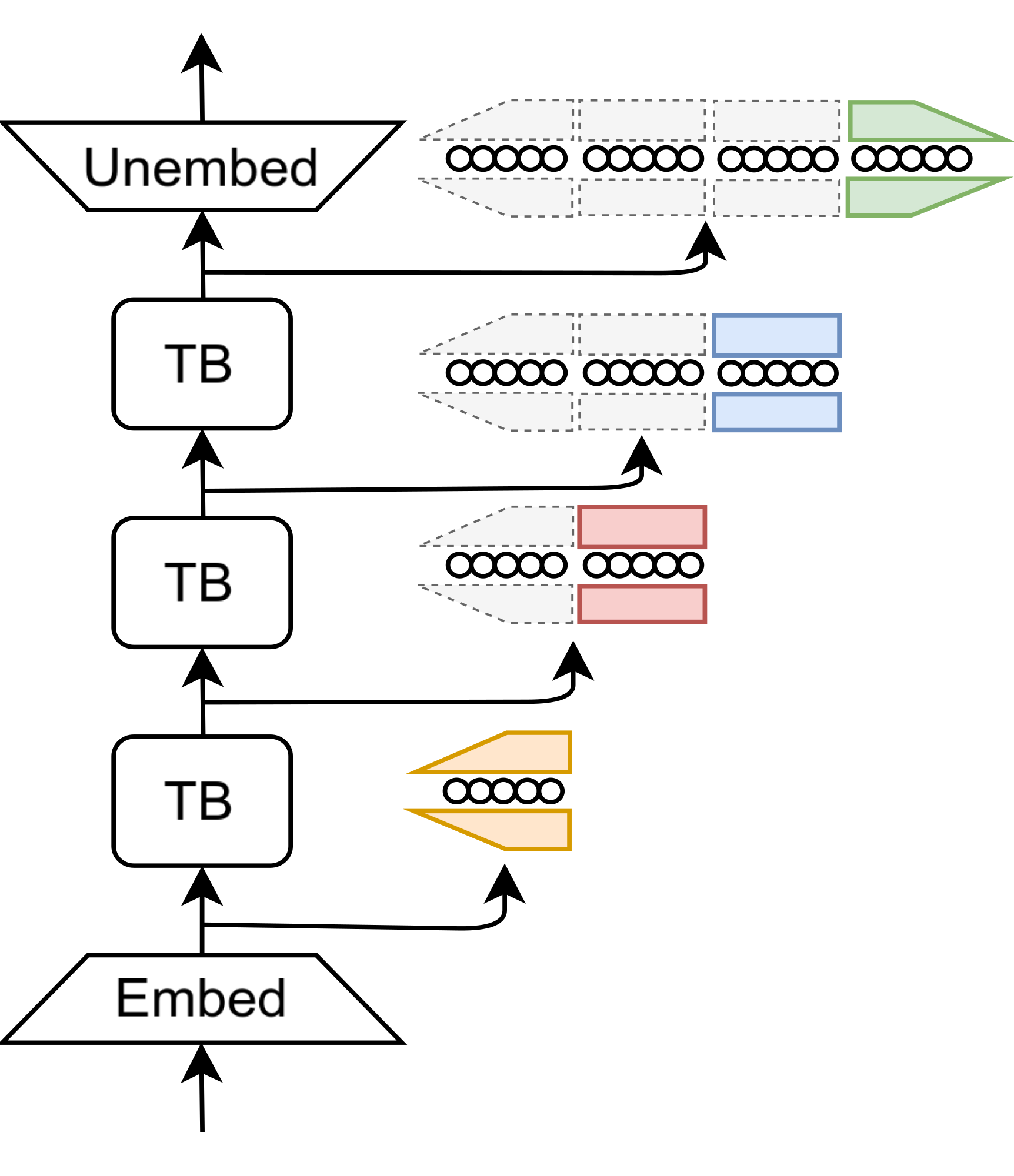}
	\caption{If chunks other than that for the current layer are not used, the staircase SAE degenerates back to a standard SAE.}
	\label{fig:staircase_detach_ablation}
\end{figure}

As can be seen in \Cref{fig:l0_per_chunk}, the detached gradient variant focuses entirely on the features of
the current chunk, and does not use features from previous chunks, essentially degenerating back to a standard SAE (\Cref{fig:staircase_detach_ablation}).
This variant underperformed both staircase SAEs as well as standard SAEs, indicating that it is important to allow
the optimizer to optimize all weights in each layer to minimize reconstruction loss.

\FloatBarrier

\section{DynamicTanh} 
\label{app:dyt}
Normalization layers are a cornerstone of modern deep learning, widely considered essential
for stable and efficient training \cite{ba2016layernorm}. 
The paper ``Transformers without Normalization'' \cite{zhu2025transformers} 
challenges this convention by introducing DynamicTanh (DyT), a simple 
element-wise operation proposed as a drop-in replacement
for normalization layers in Transformers.

\begin{equation}
    \text{DyT}(x) = \tanh(\alpha x) \odot \gamma + \beta
    \label{eq:dyt}
\end{equation}
where:
\begin{itemize}
    \item $x \in \mathbb{R}^{d_{\text{model}}}$ is the input tensor.
    \item $\alpha \in \mathbb{R}^{d_{\text{model}}}$ is a learnable scaling factor
    for each element of $x$.
    \item $\gamma, \beta \in \mathbb{R}^{d_{\text{model}}}$ are the standard scale and shift parameters
    of a standard normalization layer.
\end{itemize}

Compare with LN:
\begin{equation}
    \text{LN}(x) = \frac{x - \mathbb{E}[x]}{\sqrt{\mathbb{V}[x] + \epsilon}} \odot \gamma + \beta
    \label{eq:ln}
\end{equation}
where $\mathbb{E}[x]$ and $\mathbb{V}[x]$ are the emperical mean and variance of the input tensor $x$ across
the batch and sequence dimensions, and both $\gamma$ and $\beta$ are learnable parameters as in DyT.

The inspiration for DyT stems from the observation that Layer Normalization (LN) in
Transformers often produces input-output mappings that are sigmoid-shaped and resemble a $\tanh$
function, particularly in deeper layers (see Figure 2 in~\cite{zhu2025transformers}). They demonstrate that by incorporating DyT, Transformers
without normalization can match or even exceed the performance of their normalized
counterparts, suggesting that the primary
beneficial effect of normalization in these architectures might be the adaptive non-linear
squashing of activations, rather than strict statistical normalization.

The benefit to us of DyT is that this function is computed element-wise, which allows
us to effectively compute the Jacobian of the MLP block when LayerNorm is replaced with 
DynamicTanh. If we were to just use LayerNorm, this calculation would be too expensive (\Cref{app:jacobian_derivation_ln}).

\newcommand{\hatx}{\vec{\hat{x}}}
\newcommand{\haty}{\vec{\hat{y}}}
\newcommand{\dsae}{d_{\text{SAE}}}
\newcommand{\dmodel}{d_{\text{model}}}
\section{Jacobian Derivation}
\label{app:jacobian_derivation}

The following description of the Jacobian derivation for JSAEs across an MLP layer 
is taken directly from \cite{farnik2025jacobian}, where one can find a detailed derivation.

We wish to compute the Jacobian of the function $f_s = \ency \circ f \circ \decx \circ \topk$,
describing the mapping $\sy = f_s(\sx)$ from the hidden latents $\sx$ of a TopK SAE before a MLP layer, 
to the hidden latents $\sy$ of a TopK SAE 
after the MLP layer. Here, $\topk$ denotes the TopK activation function, 
$\hatx = \decx(\sx) = \Wdecx \sx + \bdecx$ is the decoder of the first SAE, 
$\sy = \ency(\y) = \topk\left( \Wency \y + \bency \right)$ is the encoder of the second SAE, and 
$\y = f(\hatx) = \Wout \phi_{\text{MLP}}(\vec{z}) + \vec{b}_\text{2}$ with $\vec{z} = \Win \hatx + \vec{b}_\text{1}$
is the MLP.

\cite{farnik2025jacobian} show that the Jacobian of $f_s = \ency \circ f \circ \decx \circ \topk$ 
can be computed efficiently as
\begin{align}
    J_{f_s,ij} &=
    \begin{cases}
        \sum_{k\ell m}
        \Wencyik \,
        W_{\text{2},k\ell} \,
        \phi_{\text{MLP}}'(z_\ell) \,
        W_{\text{1},\ell m} \,
        \Wdecxmj
        & \text{if } i\in\mathcal{K}_\text{2}\land j \in \mathcal{K}_\text{1} \\
        0 & \text{otherwise}
    \end{cases}
\end{align}
where $\mathcal{K}_\text{1}$ and $\mathcal{K}_\text{2}$ are the sets of indices corresponding to the features
selected by the TopK activation function of the first and second SAEs respectively, 
and that the (at most) $k \times k$ non-zero elements of the Jacobian 
can be compactly represented as
\begin{align}
    \mat{J}_{f_s}^\text{(active)} &= \underbrace{\Wencyactive\W_{\text{2}}}_{\mathbb{R}^{k \times d_{\text{MLP}}}} \cdot 
    \underbrace{\phi_{\text{MLP}}'(\vec{z})}_{\mathbb{R}^{d_{\text{MLP}} \times d_{\text{MLP}}}} \cdot
    \underbrace{\W_\text{1} \Wdecxactive}_{\mathbb{R}^{d_{\text{MLP}} \times k}}
\end{align}
where $\Wencyactive \in \mathbb{R}^{k \times \dimy}$ 
and $\Wdecxactive \in \mathbb{R}^{\dimx \times k}$ contain the active rows and columns, 
i.e., the rows and columns corresponding to the $\mathcal{K}_\text{2}$ or $\mathcal{K}_\text{1}$ indices respectively.

\subsection{Jacobian Derivation for MLP Block with Skip Connection}
\label{app:jacobian_mlp_block}
We extend this by giving an efficient implementation of the Jacobian of
$g_s = \ency \circ g \circ \decx \circ \topk$, where 
\begin{equation}
	g(\vec{\hat{x}}) = \vec{\hat{x}} + \text{FF}(h(\vec{\hat{x}}))
\end{equation}
and $h$ is an element-wise normalization function (we use $h = \text{DyT}$ in practice).
\textbf{Note that we ignore the biases of the SAEs and the FF layer, as they do not change the Jacobian calculation.}

The components are:
\begin{itemize}
    \item $\sx \in \Reals^{\dsae}$ is the $k$-sparse vector of active features from the hidden layer of the input \SAE.
    \item $\barsx \in \Reals^{\dsae}$ is the result of passing $\sx$ through the TopK activation function $\tau_k$. This doesn't
    change anything $(\barsx = \sx)$, but it makes calculation of the Jacobian efficient.
    \item $\Wdecx \in \Reals^{\dsae \times \dmodel}$ is the full decoder matrix of the input \SAE.
    \item $\Wdecxactive \in \Reals^{k \times \dmodel}$ consists of the $k$ rows of $\Wdecx$ corresponding to the active features in $\sx$.
    \item The dense vector $\x \in \Reals^{\dmodel}$ is reconstructed from $\sx$ as $\x = (\Wdecxactive)^T \sx$. The decoder bias $\bdecx$ is ignored for Jacobian calculation.
    \item $h: \Reals^{\dmodel} \to \Reals^{\dmodel}$ is an element-wise normalization function, that is, there
	is some function $\psi: \Reals \to \Reals$ such that $h(\x)_i = \psi(x_i)$.
	Its Jacobian $\frac{\partial h(\x)_i}{\partial x_j}$ 
	is $\text{diag}(h'(\x))$, a diagonal matrix with elements $\psi'(x_i)$ on its diagonal. 
	As an abuse of notation, we use $h(x_i)$ in place of $\psi(x_i)$.
    \item The MLP (or \FFN) processes $\vec{a} = h(\x)$. It is defined (ignoring biases) as 
	$$
	\text{MLP}(\vec{a}) = \Wout \phi_{\text{MLP}}(\Win \vec{a}).
	$$
    \begin{itemize}
        \item $\Win \in \Reals^{d_{\text{mlp}} \times \dmodel}$ is the MLP input weight matrix.
        \item $\Wout \in \Reals^{\dmodel \times d_{\text{mlp}}}$ is the MLP output weight matrix.
        \item $\phi_{\text{MLP}}$ is the MLP's element-wise activation function. Its Jacobian with respect to its input $\vec{z}_1 = \Win \vec{a}$ is 
		$\text{diag}(\phi'_{\text{MLP}}(\vec{z}_1))$.
    \end{itemize}
    \item $\Wency \in \Reals^{\dmodel \times \dsae}$ is the full encoder matrix of the output \SAE.
    \item $\Wencyactive \in \Reals^{\dmodel \times k}$ consists of the $k$ columns of $\Wency$ corresponding to the active features that will form $\sy$. 
    \item $\sy \in \Reals^{\dsae}$ is the $k$-sparse vector of output features for the output \SAE.
\end{itemize}

We can decompose the expression into two paths: the skip connection, and the MLP path, given that all of the functions
are linear, or act element-wise. 
\begin{align*}
g_s(\sx) &= (\ency \circ g \circ \decx \circ \topk)(\sx) \\
&= (\ency \circ g \circ \decx)(\barsx) \\
&= (\ency \circ g)(\decx(\barsx)) \\
&= \ency(\decx(\barsx) + \text{FF}(\text{DyT}(\decx(\barsx)))) \\
&= \ency(\decx(\barsx)) + \ency(\text{FF}(\text{DyT}(\decx(\barsx)))) \\
&= \underbrace{(\ency \circ \decx \circ \topk)(\sx)}_{\sy^{\text{skip}}} + \underbrace{(\ency \circ \text{FF} \circ \text{DyT} \circ \decx \circ \topk)(\sx)}_{\sy^{\text{FFh}}} \\
&= g_s^{\text{skip}}(\sx) + g_s^{\text{FFh}}(\sx)
\end{align*}
with $\sy^{\text{skip}} = g_s^{\text{skip}}(\sx) = (\ency \circ \decx \circ \topk)(\sx)$
and $\sy^{\text{FFh}} = g_s^{\text{FFh}}(\sx) = (\ency \circ \text{FF} \circ \text{DyT} \circ \decx \circ \topk)(\sx)$.
We can thereby compute the Jacobian as the sum of the Jacobians of the two paths.
\begin{align}
\mat{J}_{g_s} &= \mat{J}_{g_s}^{\text{skip}} + \mat{J}_{g_s}^{\text{FFh}}
\end{align}

\subsubsection{Skip Path Jacobian (\texorpdfstring{$\mat{J}_{\text{skip}}$}{J\_skip})}
The skip path Jacobian is the Jacobian of the skip path, which is a linear function (if focused on the features chosen by the TopK activation).

\begin{equation}
	\sx \overset{\tau_x}{\longrightarrow} \barsx \overset{\decx}{\longrightarrow} \hatx 
	\overset{\ency}{\longrightarrow} \sy^{\text{skip}} 	
\end{equation}

\newcommand{\sskip}{\mathbf{s}^\text{skip}}

\begin{equation}
	\mat{J}_{g_s, ij}^{\text{skip}} = \frac{\partial s^\text{skip}_{y,i}}{\partial s_{x,j}} = 
\sum_{pq} \frac{\partial s^\text{skip}_{y,i}}{\partial \hat{x}_p} \frac{\partial \hat{x}_p}{\partial \bar{s}_{q}} \frac{\partial \bar{s}_{q}}{\partial s_{x,j}}
\end{equation}
Computing each term:

\begin{align}
\frac{\partial s^\text{skip}_{y,i}}{\partial \hat{x}_p} &= \frac{\partial \tau_k(\Wency \hat{x})_i}{\partial \hat{x}_p} 
=  \W^\text{enc}_{y, ip} \llbracket i \in \mathcal{K}_2 \rrbracket \\
\frac{\partial \hat{x}_p}{\partial \bar{s}_{q}} &= \frac{\partial (\Wdecx \bar{s})_p}{\partial \bar{s}_q} = \W^\text{dec}_{x, pq} \\
\frac{\partial \bar{s}_{q}}{\partial s_{x,j}} &= \frac{\partial \tau_k(s_{x,q})}{\partial s_{x,j}} = \delta_{qj} \llbracket j \in \mathcal{K}_1 \rrbracket 
\end{align}
which, when combined gives:
\begin{align}
\mat{J}_{g_s, ij}^{\text{skip}} &= \sum_{p,q} \W^\text{enc}_{y, ip} \W^\text{dec}_{x, pq} \delta_{qj} \llbracket i \in \mathcal{K}_1 \rrbracket 
\llbracket j \in \mathcal{K}_1 \rrbracket \\
&= \sum_{p} \W^\text{enc}_{y, ip} \W^\text{dec}_{x, pj} \llbracket i \in \mathcal{K}_1 \rrbracket \llbracket j \in \mathcal{K}_1 \rrbracket \\
&= \W^\text{enc(active)}_y \W^\text{dec(active)}_x
\end{align}

\newcommand{\dmlp}{d_{\text{mlp}}}

\subsection{MLP Path Jacobian (\texorpdfstring{$\mat{J}_{\text{MLP}}$}{J\_MLP})}
Let $\sy^{\text{FFh}}$ be the sparse output contribution from the MLP path. 
The path through the FF Block is computed as
\begin{equation}
	\sx \overset{\tau_x}{\longrightarrow} \barsx \overset{\decx}{\longrightarrow} \hatx \overset{h}{\longrightarrow} h(\hatx) 
	\overset{\W_1}{\longrightarrow} \vec{z} \overset{\phi}{\longrightarrow} \phi_{\text{MLP}}(\vec{z}) 
	\overset{\W_2}{\longrightarrow} \y \overset{\ency}{\longrightarrow} \sy^{\text{FFh}} 	
\end{equation}
\begin{align}
\mat{J}_{g_s,ij}^{\text{FFh}} &= \frac{\partial s_{y,i}^{\text{FFh}}}{\partial s_{x,j}}
= \sum_{pqrtuv} \frac{\partial s_{y,i}^{\text{FFh}}}{\partial \y_p}
\frac{\partial \y_p}{\partial \phi_{\text{MLP}}(\vec{z})_q}
\frac{\partial \phi_{\text{MLP}}(\vec{z})_q}{\partial \vec{z}_r}
\frac{\partial \vec{z}_r}{\partial h(\hat{\x})_t}
\frac{\partial h(\hat{\x})_t}{\partial \hat{\x}_u}
\frac{\partial \hat{\x}_u}{\partial \bar{s}_v}
\frac{\partial \bar{s}_v}{\partial s_{x,j}}
\end{align}
As before, we compute each of these terms separately:
\begin{align}
	\frac{\partial s_{y,i}^{\text{FFh}}}{\partial \y_p} &= \frac{\partial \tau_k(\Wency \y)_i}{\partial \y_p} = \W^\text{enc}_{y, ip} \llbracket i \in \mathcal{K}_2 \rrbracket \\
	\frac{\partial \y_p}{\partial \phi_{\text{MLP}}(\vec{z})_q} &= \frac{\partial \W_2 \phi_{\text{MLP}}(\vec{z})_p}{\partial \phi_{\text{MLP}}(\vec{z})_q} 
	= \W_{2,pq} \\
	\frac{\partial \phi_{\text{MLP}}(\vec{z})_q}{\partial \vec{z}_r} &= \text{diag}(\phi'_{\text{MLP}}(\vec{z}))_{qr} =
	\delta_{qr} \phi'_{\text{MLP}}(z_r) \\
	\frac{\partial \vec{z}_r}{\partial h(\hat{\x})_t} &= \frac{\partial (\W_1 \hat{\x})_r}{\partial h(\hat{x})_t} = \W_{1,rt} \\
	\frac{\partial h(\hat{\x})_t}{\partial \hat{\x}_u} &= \text{diag}(h'(\hat{\x}))_{tu} = \delta_{tu} h'(\hat{x}_u) \\
	\frac{\partial \hat{\x}_u}{\partial \bar{s}_{x,v}} &= \frac{\partial (\Wdecx \bar{s}_{x})_u}{\partial \bar{s}_{x,v}} = \W^\text{dec}_{x, uv} \\
	\frac{\partial \bar{s}_{x,v}}{\partial s_{x,j}} &= \frac{\partial \tau_k(s_{x,v})}{\partial s_{x,j}} = \delta_{vj} \llbracket j \in \mathcal{K}_1 \rrbracket
\end{align}

Substituting these into the sum for $\mat{J}_{g_s,ij}^{\text{FFh}}$:
\begin{align*}
\mat{J}_{g_s,ij}^{\text{FFh}} &= \sum_{pqrtuv} 
\W^\text{enc}_{y, ip} \llbracket i \in \mathcal{K}_2 \rrbracket
\W_{2,pq}
\delta_{qr} \phi'_{\text{MLP}}(z_r)
\W_{1,rt}
\delta_{tu} h'(\hat{x}_u)
\W^\text{dec}_{x, uv}
\delta_{vj} \llbracket j \in \mathcal{K}_1 \rrbracket \\
\intertext{Dropping the terms $\delta_{qr}, \delta_{tu}, \delta_{vj}$ and replacing $q \to r, t \to u, v \to j$, we get}
&= \sum_{pru} \W^\text{enc}_{y, ip} \W_{2,pr} \W_{1,ru} \phi'_{\text{MLP}}(z_r) h'(\hat{x}_u) \W^\text{dec}_{x, uj} \llbracket i \in \mathcal{K}_2 \rrbracket \llbracket j \in \mathcal{K}_1 \rrbracket
\end{align*}

Applying these simplifications, the sum becomes:
\begin{align*}
\mat{J}_{g_s,ij}^{\text{FFh}} = \llbracket i \in \mathcal{K}_2 \rrbracket \llbracket j \in \mathcal{K}_1 \rrbracket \sum_{pru}
 (\Wency)_{ip}
 \W_{2,pr}
 \phi'_{\text{MLP}}(z_r)
 \W_{1,ru}
 h'(\hat{x}_u)
 (\Wdecx)_{uj}
\end{align*}
This sum represents the $(i,j)$-th element of a matrix product. Let $\mathbf{\nabla}^\phi = \text{diag}(\phi'_{\text{MLP}}(\vec{z}))$ be a diagonal matrix with elements $\phi'_{\text{MLP}}(z_r)$ on its diagonal, and $\mathbf{\nabla}^h = \text{diag}(h'(\hatx))$ be a diagonal matrix with elements $h'(\hat{x}_u)$ on its diagonal.
Then the $(i,j)$-th element is:
\begin{align}
\mat{J}_{g_s,ij}^{\text{FFh}} &= \left( \Wency \W_2 \mathbf{\nabla}^\phi \W_1 \mathbf{\nabla}^h \Wdecx \right)_{ij} \llbracket i \in \mathcal{K}_2 \rrbracket \llbracket j \in \mathcal{K}_1 \rrbracket
\end{align}
In matrix form, considering only the active features (where $i \in \mathcal{K}_2$ and $j \in \mathcal{K}_1$), the active block of the Jacobian is:
\begin{equation}
\mat{J}_{g_s}^{\text{FFh, (active)}} = \W^\text{enc(active)}_y \W_2 \mathbf{\nabla}^\phi \W_1 \mathbf{\nabla}^h \W^\text{dec(active)}_x
\end{equation}
Finally, the terms $\mathbf{\nabla}^\phi, \mathbf{\nabla}^h$ are diagonal matrices (stored as a vector), and so their product with an adjacent matrix is cheap
to compute:
\begin{equation}
(\mathbf{\nabla}^\phi \W)_{ij} = \sum_k \mathbf{\nabla}^\phi_{ik} \W_{kj} = \mathbf{\nabla}^\phi_{ii} \W_{ij} = \phi'_{\text{MLP}}(z_i) \W_{ij}
\end{equation}

Combining the two terms together, we get:
\begin{equation}
\mat{J}_{g_s} = \mat{J}_{g_s}^{\text{skip}} + \mat{J}_{g_s}^{\text{FFh}} = \W^\text{enc(active)}_y ( \mathbf{I} + \W_2 \mathbf{\nabla}^\phi \W_1 \mathbf{\nabla}^h) \W^\text{dec(active)}_x
\end{equation}
though for efficiency, we compute the two terms separately and add them together.

\newcommand{\dimembd}{D_{\text{embd}}}
\newcommand{\dimmlp}{D_{\text{mlp}}}

\newcommand{\WdecxK}{W_{D,\mathcal{K}_1}} %
\newcommand{\WencyK}{W_{E,\mathcal{K}_2}} %

\subsection{Jacobian with LayerNorm}
\label{app:jacobian_derivation_ln}

Consider for a moment an arbitrary activation $x \in \mathbb{R}^{s \times d}$ that flows through
a transformer: a set of $d$-dimensional activations, one for each sequence position $s$,
and some arbitrary module $\psi: \mathbb{R}^{s \times d} \to \mathbb{R}^{s \times d}$ that performs
some computation on those activations. In the most general case, the Jacobian of this module $\nabla^\psi$ is a
tensor of shape $(s, d, s, d)$, requiring $O(s^2 d^2)$ memory to store. For any reasonable size of sequence length $s$ and
model dimension $d$, this is prohibitively large. For gpt-2 small, $s = 1024$ and $d = 768$, this is $\approx 6 \cdot 10^{11}$ elements,
or about 2.25TB of memory to store.

Given assumptions about the module $\psi$, we can reduce the memory requirements for the Jacobian:
\begin{itemize}
    \item \textbf{Linear}: The Jacobian will be a constant matrix, so we can ignore the sequence dimensions $s, s'$.
    \item \textbf{Elementwise}: The Jacobian will be sparse, with only the diagonal being non-zero. Off-diagonal terms $(d,d')$ can be ignored.
    \item \textbf{Sequence Independent}: We only need to track the Jacobian with respect to one sequence position $s$. Cross-terms $(s,s')$ can be ignored.
\end{itemize}

\begin{table}[htbp]
  \centering
  \label{tab:jacobian_summary}
  \small
  \definecolor{lightgreen}{rgb}{0.7, 1.0, 0.7}
  \definecolor{lightyellow}{rgb}{1.0, 1.0, 0.7}
  \definecolor{lightorange}{rgb}{1.0, 0.8, 0.6}
  \definecolor{lightred}{rgb}{1.0, 0.7, 0.7}

  \begin{tabular}{c c c l l r l}
    \toprule
    Linear & Elementwise & Seq. Indep. & Jacobian Form & Cost & RAM & Examples \\
    \midrule
    \textcolor{green}{\checkmark} & \textcolor{green}{\checkmark} & \textcolor{green}{\checkmark}
    & $\delta_{ss'} \delta_{dd'} \nabla^\psi_{d}$
    & \cellcolor{lightgreen} $O(d)$
    & \cellcolor{lightgreen} 4kB
    & $\mathbf{\gamma} \odot \x + \mathbf{\beta}$ \\
    \midrule
    \textcolor{green}{\checkmark} & \textcolor{red}{\ding{55}} & \textcolor{green}{\checkmark}
    & $\delta_{ss'} \nabla^\psi_{dd'}$
    & \cellcolor{lightyellow} $O(d^2)$
    & \cellcolor{lightyellow} 4MB
    & $\W \x + \mathbf{b}$ \\
    \midrule
    \textcolor{red}{\ding{55}} & \textcolor{green}{\checkmark} & \textcolor{green}{\checkmark}
    & $\delta_{ss'} \delta_{dd'} \nabla^\psi_{sd}$
    & \cellcolor{lightyellow} $O(sd)$
    & \cellcolor{lightyellow} 3MB
    & GeLU, ReLU, DyT \\
    \midrule
    \textcolor{red}{\ding{55}} & \textcolor{red}{\ding{55}} & \textcolor{green}{\checkmark}
    & $\delta_{ss'} \nabla^\psi_{sd'd}$
    & \cellcolor{lightorange} $O(sd^2)$ 
    & \cellcolor{lightorange} 768MB
    & LayerNorm \\
    \midrule
    \textcolor{red}{\ding{55}} & \textcolor{red}{\ding{55}} & \textcolor{red}{\ding{55}}
    & $\nabla^\psi_{sds'd'}$
    & \cellcolor{lightred} $O(s^2 d^2)$
    & \cellcolor{lightred} 2.25TB
    & Self-Attention \\
    \bottomrule
  \end{tabular}
  \caption{Summary of Jacobian structures and memory requirements for a
  module $\psi: \mathbb{R}^{s \times d} \to \mathbb{R}^{s \times d}$, based on assumptions of linearity, 
  elementwise-ness, and sequence independence. Memory costs estimated for GPT-2 Small.}
\end{table}

LayerNorm may be sequence independent, but being neither linear nor elementwise, the Jacobian is rather large, making intractible to evaluate
inside a tight training loop. This motivates the use of replacing it with an elementwise normalization operation such as 
DynamicTanh \cite{zhu2025transformers} or fine-tuning to remove the LayerNorm entirely \cite{heimersheim2024removeln}

\section{Experimental Comparison Setup}
\label{app:staircase_variants}
We compared the proposed Staircase SAE against several baseline configurations.
All experiments described here used TopK SAEs with $K=10$. The expansion
factor of the SAE hidden layer relative to the model's activation dimension
($d_{\text{model}}$) is specified for each model.

The models compared are:
\begin{itemize}
	\item \textbf{Top$k$-x8}: Standard TopK SAEs trained independently per layer. The
	      hidden dimension $d_{\text{sae}}$ was set to
	      $8 \times d_{\text{model}}$ for all layers.
	\item \textbf{Top$k$-x40}: Identical to \emph{Top$k$-x8}, but with a hidden dimension of
	      $40 \times d_{\text{model}}$ for all layers.
	\item \textbf{Top$k$-x40-tied}: Identical to \emph{Top$k$-x40}, but encoder
	and decoder matrices are shared across all layers.
	\item \textbf{Staircase-x8}: The proposed Staircase SAE
	      architecture using TopK activation, as described in \Cref{subsec:staircase_sae}.
	\item \textbf{Staircase-untied-x8}: Same as \emph{Staircase-x8}, but with
	      independent encoder and decoder matrices for each layer.
	\item \textbf{Staircase-Detach}: A variant of \emph{Staircase-x8} where each layer $i$ only
	      optimizes parameters related to its own feature chunk $i$. As
	      mentioned, this approach performed poorly (\Cref{sec:staircase_sae}).
\end{itemize}

We note that the first four models (Staircase-Detach, Top$k$-x8,
Staircase-x8, Top$k$-x40-tied) have approximately the
same number of trainable parameters (ignoring the biases $\mathbf{b}_{enc}^i, \mathbf{b}_{dec}^i$,
which contribute negligibly to the total count compared to the weight
matrices). The last two models (Staircase-untied-x8 and Top$k$-x40) have
significantly more parameters and, unsurprisingly, tend to achieve better
reconstruction performance, serving as benchmarks for capacity.

\begin{figure}[htbp]
	\centering
	\includegraphics[width=\linewidth]{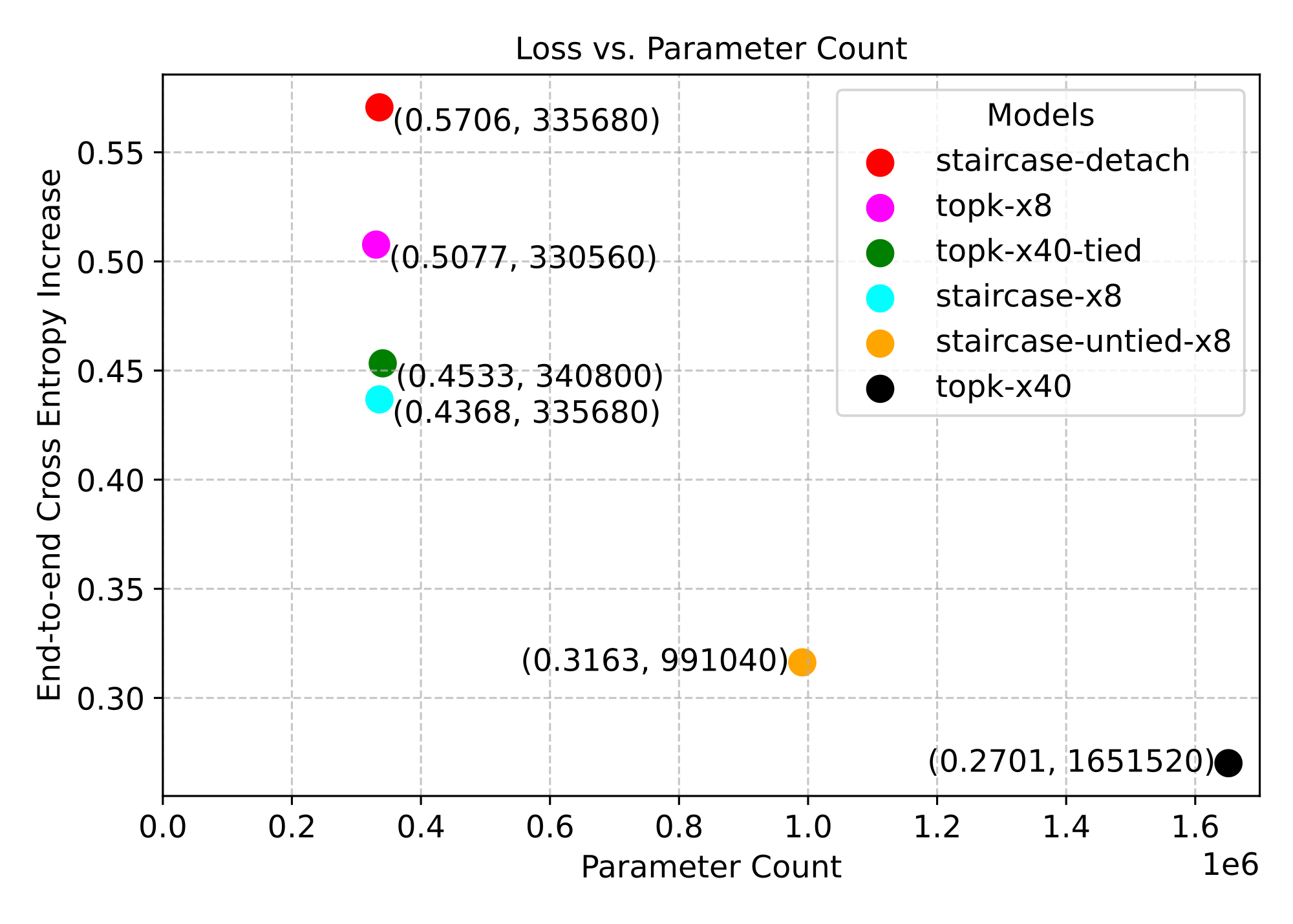}
	\caption{For a given parameter count, the staircase SAE (\textbf{staircase-x8}) has a lower increase in
	CE loss over the validation set than the baseline SAEs described in \Cref{app:staircase_variants}.}
	\label{fig:staircase_ce_increase}
\end{figure}

\section{Scoring Connections}
\subsection{Integrated Gradient Attribution}\label{IGA}
Here we give a more detailed description of integrated gradient attribution, and describe the details of the method we use to implement it.

Integrated Gradient Attribution is a solution to the local attributions problem. That is, given a vector $v$, and a real valued differentiable function $f$ this method produces an attribution $a_i$ for each coordinate of $v$ satisfying
$$f(v) = \sum_i a_i.$$
To compute gradient attribution, choose a base point $b$ in the same vector space as $v$. Each $a_i$ is determined by the formula
$$a_i = (v-b)\cdot \int_0^1 \partial_i f(b(1-z) + zv) dz,$$
it is the integral of $\partial_i$ of $f$ along a straight line from $b$ to $v$.

To score cross-SAE latent connections, we repeat this process once for every downstream latent. For the function $f$, we decode latents in SAE$_0$, pass them through the model, and encode them to latents in SAE$_1$.

We always chose the base point $b$ to be the origin. For the data in this paper, we approximated the integral as a Riemann sum with 5 terms.

Since integrated gradients only give local data, we determined global connection scores by sampling latent activations from $576$ points in the training data, and using the root mean square of every local attribution as the overall connection score.

This process was extremely consistent, computing scores twice with different random samples of data resulted in virtually identical scores.

\FloatBarrier
\section{Jacobian coefficient tuning}
\label{app:jacobian_coeff_tuning}

At each block, a JSAE pair was trained with a Jacobian coefficient $\lambda$ taking geometrically spaced values in the set: 
\begin{align}\label{equ:Jacobian_coefficient_set}
    \{0, 1, 1.2, 1.5, 1.8, 2.2, 2.7, 3.3, 3.9, 4.7, 5.6, 6.8, 10\} \times 10^{-3}.
\end{align}
following the E12 series \cite{wikipediaEseries} of preferred numbers, together with $\lambda = 0$.
Recall that for $\lambda = 0$, the JSAE pair degenerates to a TopK SAE pair. 
For each such pair, an ablation curve was produced and the corresponding SCALAR score was calculated, see Table \ref{tab:Jacobian_coefficient_SCALAR}.
We also measured the CE loss increase when the JSAEs were spliced in over a large range of $\lambda$, and then swept over the critical range
where it started to non-trivially impact the performance of reconstruction.

\begin{figure}[htbp]
	\centering
	\begin{subfigure}[b]{\textwidth}
		\centering
		\includegraphics[width=\textwidth]{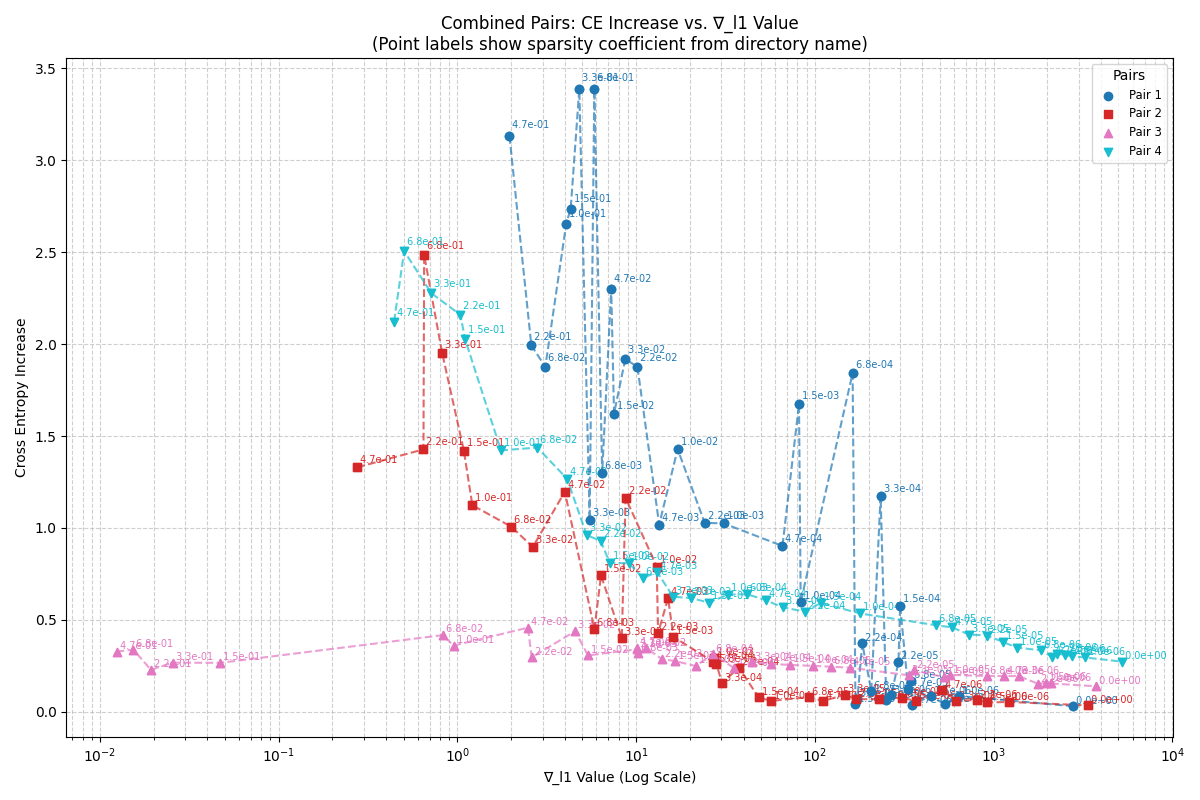}
		\label{fig:pareto_all_layer}
	\end{subfigure}

	\begin{subfigure}[b]{\textwidth}
		\centering
		\includegraphics[width=\textwidth]{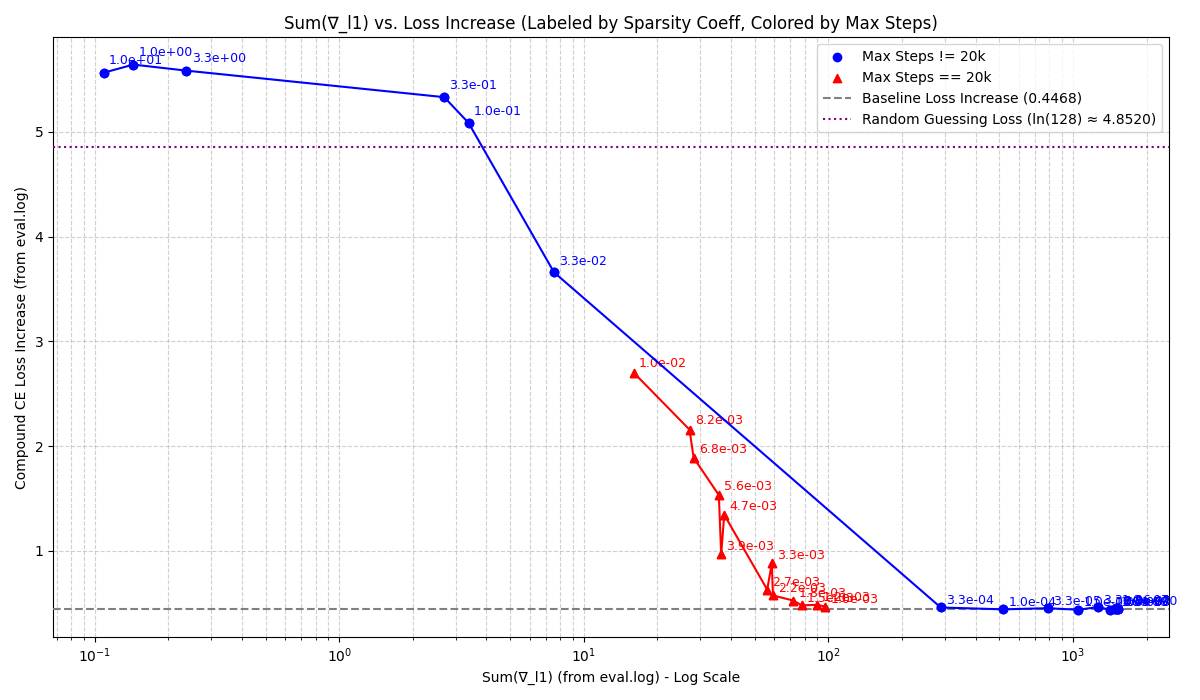}
		\label{fig:pareto_sweep}
	\end{subfigure}

	\caption{Pareto curves showing the trade-off between reconstruction performance and Jacobian coefficient values. 
	The top image shows overall performance across all layers and the bottom image shows 
	the concentrated sweep over the sparsity coefficient $\lambda$ discussed in \Cref{app:jacobian_coeff_tuning}.}
	\label{fig:pareto_curves_page1}
\end{figure}
\begin{figure}[htbp]
	\centering
	\begin{subfigure}[b]{0.456\textwidth}
		\centering
		\includegraphics[height=0.95\textwidth, angle=90]{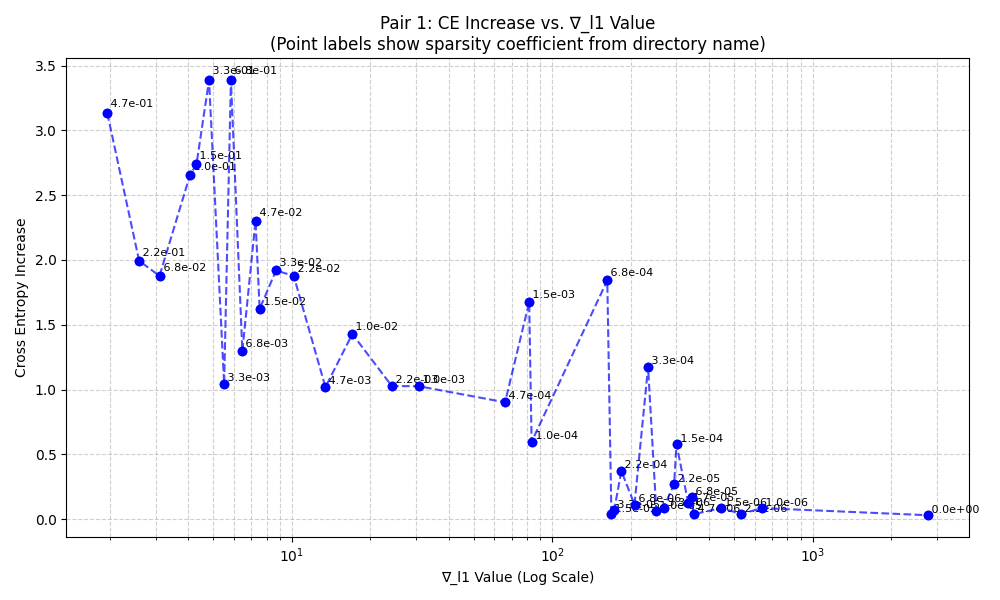}
		\label{fig:pareto_layer1}
	\end{subfigure}
	\hfill
	\begin{subfigure}[b]{0.456\textwidth}
		\centering
		\includegraphics[height=0.95\textwidth, angle=90]{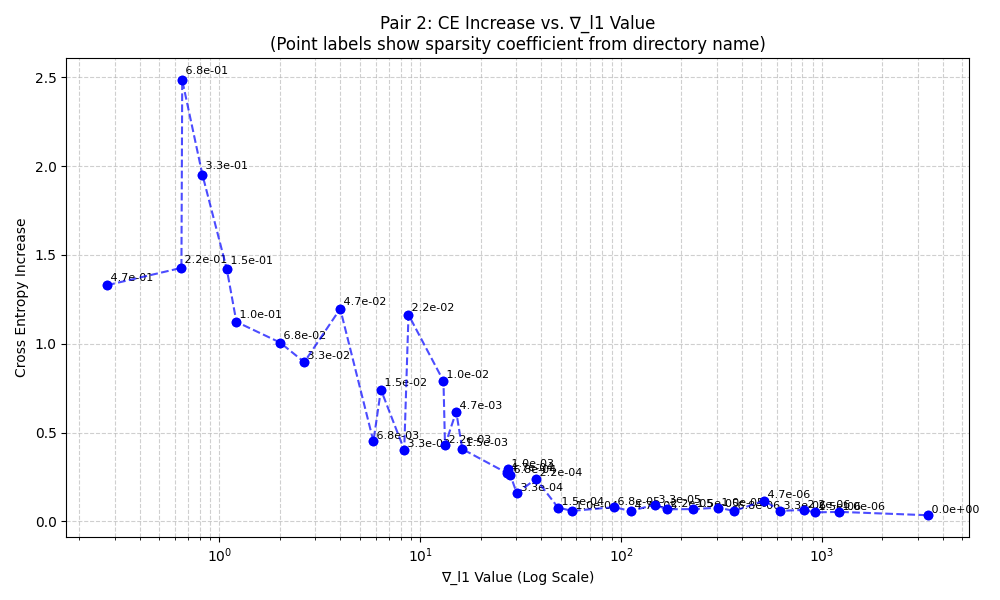}
		\label{fig:pareto_layer2}
	\end{subfigure}
	
	\begin{subfigure}[b]{0.456\textwidth}
		\centering
		\includegraphics[height=0.95\textwidth, angle=90]{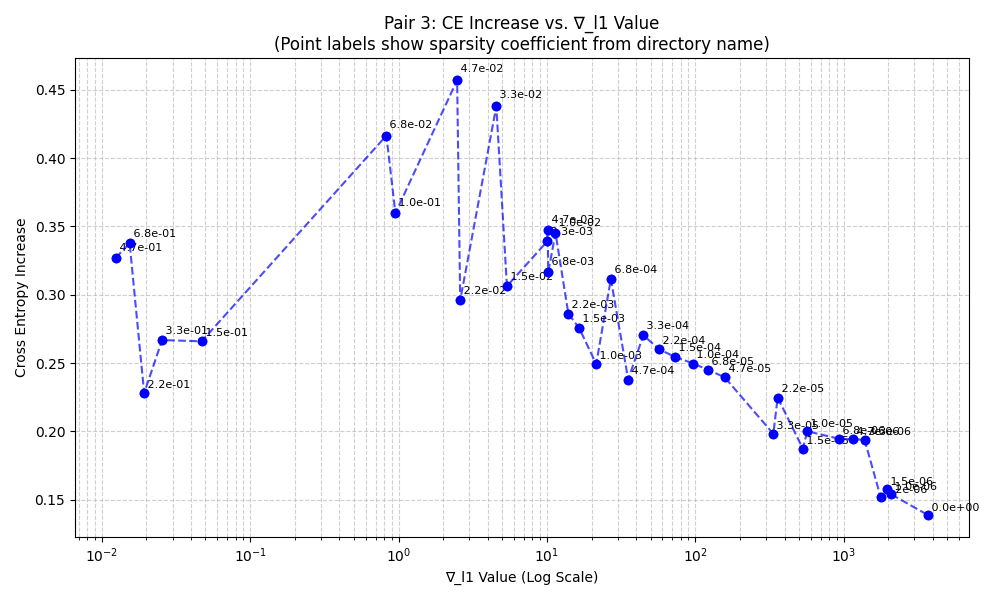}
		\label{fig:pareto_layer3}
	\end{subfigure}
	\hfill
	\begin{subfigure}[b]{0.456\textwidth}
		\centering
		\includegraphics[height=0.95\textwidth, angle=90]{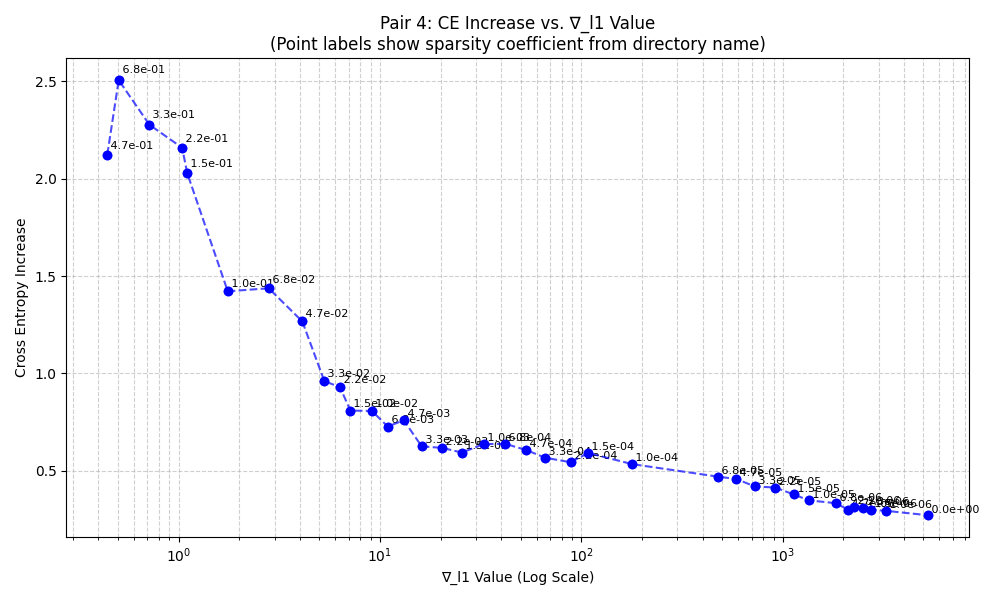}
		\label{fig:pareto_layer4}
	\end{subfigure}
	
	\caption{Pareto curves showing the trade-off between reconstruction performance and 
	Jacobian coefficient values for individual layers. 
	Arranged in a 2×2 grid: layer 1 (top left), layer 2 (top right), 
	layer 3 (bottom left), and layer 4 (bottom right).}
	\label{fig:pareto_curves_page2}
\end{figure}

To ensure maximum competitiveness between TopK and JSAE variants, the JSAE results reported in Section \ref{sec:results} 
take the (non-zero) Jacobian coefficient corresponding to the minimum SCALAR score for each layer.

\begin{table}[h]
    \centering
    \begin{tabular}{lccccc}
        \toprule
        & \textbf{Layer 0} ($10^{4}$) & \textbf{Layer 1} ($10^{5}$) & \textbf{Layer 2} ($10^{5}$) & \textbf{Layer 3} ($10^{5}$)\\
        \midrule
        \textbf{0}  & $ 0.098 \pm 0.007 $ & $ 0.555 \pm 0.028 $ & $ 1.569 \pm 0.062 $ & $ 1.347 \pm 0.060 $\\
        \textbf{1}  & $ 0.733 \pm 0.049 $ & $ 0.483 \pm 0.025 $ & $ 1.540 \pm 0.063 $ & $ 1.544 \pm 0.068 $\\
        \textbf{1.2}  & $ 0.927 \pm 0.082 $ & $ \bf{0.419 \pm 0.022} $ & $ 1.524 \pm 0.062 $ & $ 1.591 \pm 0.069 $\\
        \textbf{1.5}  & $ 0.874 \pm 0.057 $ & $ 0.436 \pm 0.023 $ & $ 1.540 \pm 0.063 $ & $ 1.545 \pm 0.066 $\\
        \textbf{1.8}  & $ \bf{0.686 \pm 0.054} $ & $ 0.474 \pm 0.025 $ & $ 1.499 \pm 0.061 $ & $ 1.552 \pm 0.066 $\\
        \textbf{2.2}  & $ 1.003 \pm 0.071 $ & $ 0.613 \pm 0.033 $ & $ 1.505 \pm 0.063 $ & $ \bf{1.536 \pm 0.065} $\\
        \textbf{2.7}  & $ 1.154 \pm 0.082 $ & $ 0.638 \pm 0.034 $ & $ 1.445 \pm 0.060 $ & $ 1.635 \pm 0.068 $\\
        \textbf{3.3}  & $ 1.340 \pm 0.091 $ & $ 1.151 \pm 0.053 $ & $ \bf{1.416 \pm 0.060} $ & $ 1.702 \pm 0.071 $\\
        \textbf{3.9}  & $ 1.148 \pm 0.079 $ & $ 1.285 \pm 0.057 $ & $ 2.009 \pm 0.079 $ & $ 1.633 \pm 0.066 $\\
        \textbf{4.7}  & $ 1.270 \pm 0.073 $ & $ 1.654 \pm 0.070 $ & $ 2.490 \pm 0.094 $ & $ 1.777 \pm 0.069 $\\
        \textbf{5.6}  & $ 1.498 \pm 0.111 $ & $ 1.941 \pm 0.078 $ & $ 2.798 \pm 0.096 $ & $ 1.906 \pm 0.074 $\\
        \textbf{6.8}  & $ 1.455 \pm 0.083 $ & $ 2.286 \pm 0.089 $ & $ 3.010 \pm 0.104 $ & $ 1.937 \pm 0.074 $\\
        \textbf{10}  & $ 1.795 \pm 0.111 $ & $ 2.645 \pm 0.096 $ & $ 4.027 \pm 0.120 $ & $ 2.108 \pm 0.077 $\\
        \bottomrule
    \end{tabular}
    \vspace{0.5em}
	\caption{SCALAR scores for JSAE pairs with different Jacobian coefficient values (expressed in units of $10^{-3}$) across feedforward layers. The best performing non-zero Jacobian coefficient values for each layer are bolded. We observe that the optimal Jacobian coefficient varies across layers, with non-zero Jacobian coefficient values outperforming TopK (zero Jacobian coefficient) at layers one and two only.}
        \label{tab:Jacobian_coefficient_SCALAR}
\end{table}

Given that JSAE pairs are explicitly trained to reduce computational sparsity, we expect them to outperform TopK SAE pairs on our metric across all layers. Across the range of Jacobian coefficient values analysed, we find that JSAEs outperform TopK SAEs at Layer 1 and Layer 2 only. We offer two possible explanations for this. First, the approximately SCALAR-minimising Jacobian coefficient is not in the set (\ref{equ:Jacobian_coefficient_set}). This seems likely from the Layer 0 results, given the significant jump from $\lambda = 0$ to $\lambda = 10^{-3}$. Alternatively, layers 0 and 3 are less amenable to sparsification via the Jacobian. This seems plausible from the Layer 3 results where increasing the Jacobian coefficient typically results in a larger SCALAR score.

\FloatBarrier
\section{Pure Computational Sparsity Metric}
\label{app:PCSM}

Recall that the SCALAR score is computed from ablation plots produced by evaluating the KL divergence between the logits produced by the full model and the logits produced by a collection of subcircuits, see Section \ref{subsec:SCALAR} for further details. Here, we consider a slight variation of this score by instead producing ablation plots by computing the KL divergence between logits produced by the \textit{full circuit} and the logits produced by a collection of subcircuits (the same collection as above). Accordingly, this metric incorporates the
reconstruction quality of SAE pairs into the target distribution (the full circuit logits), 
which means that the resulting area under the ablation curve reflects the computational impact of 
ablating edges to a much greater degree.

\begin{table}[h]
    \centering
    \begin{tabular}{lccccc}
        \toprule
        & \textbf{Layer 0} ($10^{4}$) & \textbf{Layer 1} ($10^{5}$) & \textbf{Layer 2} ($10^{5}$) & \textbf{Layer 3} ($10^{5}$)\\
        \midrule
        \textbf{0}  & $ 0.851 \pm 0.064 $ & $ 0.422 \pm 0.020 $ & $ 1.027 \pm 0.040 $ & $ 0.386 \pm 0.018 $\\
        \textbf{1}  & $ 1.203 \pm 0.065 $ & $ 0.286 \pm 0.014 $ & $ 1.093 \pm 0.044 $ & $ 0.608 \pm 0.026 $\\
        \textbf{1.2}  & $ 1.113 \pm 0.072 $ & $ 0.220 \pm 0.010 $ & $ 1.054 \pm 0.042 $ & $ 0.628 \pm 0.027 $\\
        \textbf{1.5}  & $ 1.317 \pm 0.078 $ & $ 0.188 \pm 0.009 $ & $ 1.069 \pm 0.042 $ & $ 0.665 \pm 0.030 $\\
        \textbf{1.8}  & $ 0.887 \pm 0.059 $ & $ 0.134 \pm 0.006 $ & $ 1.039 \pm 0.041 $ & $ 0.632 \pm 0.026 $\\
        \textbf{2.2}  & $ 1.213 \pm 0.073 $ & $ 0.178 \pm 0.008 $ & $ 0.968 \pm 0.040 $ & $ 0.671 \pm 0.029 $\\
        \textbf{2.7}  & $ 0.797 \pm 0.053 $ & $ 0.081 \pm 0.004 $ & $ 0.836 \pm 0.033 $ & $ 0.708 \pm 0.029 $\\
        \textbf{3.3}  & $ 1.187 \pm 0.070 $ & $ 0.136 \pm 0.005 $ & $ 0.707 \pm 0.027 $ & $ 0.761 \pm 0.032 $\\
        \textbf{3.9}  & $ 0.635 \pm 0.039 $ & $ 0.114 \pm 0.004 $ & $ 0.582 \pm 0.018 $ & $ 0.770 \pm 0.032 $\\
        \textbf{4.7}  & $ 0.354 \pm 0.020 $ & $ 0.187 \pm 0.005 $ & $ 0.253 \pm 0.008 $ & $ 0.869 \pm 0.035 $\\
        \textbf{5.6}  & $ 0.773 \pm 0.044 $ & $ 0.160 \pm 0.005 $ & $ 0.184 \pm 0.006 $ & $ 0.965 \pm 0.035 $\\
        \textbf{6.8}  & $ 0.199 \pm 0.011 $ & $ 0.116 \pm 0.003 $ & $ 0.130 \pm 0.005 $ & $ 0.944 \pm 0.034 $\\
        \textbf{10}  & $ 0.272 \pm 0.017 $ & $ 0.142 \pm 0.003 $ & $ 0.095 \pm 0.002 $ & $ 0.563 \pm 0.020 $\\
        \bottomrule
    \end{tabular}
    \vspace{0.5em}
	\caption{Computational sparsity metric for JSAE pairs with different Jacobian coefficient values (expressed in units of $10^{-3}$) across feedforward layers. For layers 1 and 2, there exists a range of Jacobian coefficient values where we observe a trend of decreasing metric with increasing Jacobian coefficient.}
        \label{tab:Jacobian_coefficient_SCALAR_pure_comp}
\end{table}

Examining Table \ref{tab:Jacobian_coefficient_SCALAR_pure_comp}, we expect that by increasing the Jacobian coefficient the corresponding computational sparsity metric decreases. We find this is approximately the case in Layer 1, up to $\lambda = 2.7\times 10^{-3}$, and for Layer 2 across all Jacobian coefficient values. As in Appendix \ref{app:jacobian_coeff_tuning}, we suggest that this is likely due to the approximately metric-minimising Jacobian coefficient not being an element of the set (\ref{equ:Jacobian_coefficient_set}).

\FloatBarrier
\section{Ablation curves}
\label{app:ablation_curves}

\subsection{Mixed Performance Examples}
\label{app:additional_ablation_curves}

\begin{figure}[htbp]
    \centering
    \begin{subfigure}[b]{0.34125\textwidth}
        \centering
        \includegraphics[width=\textwidth]{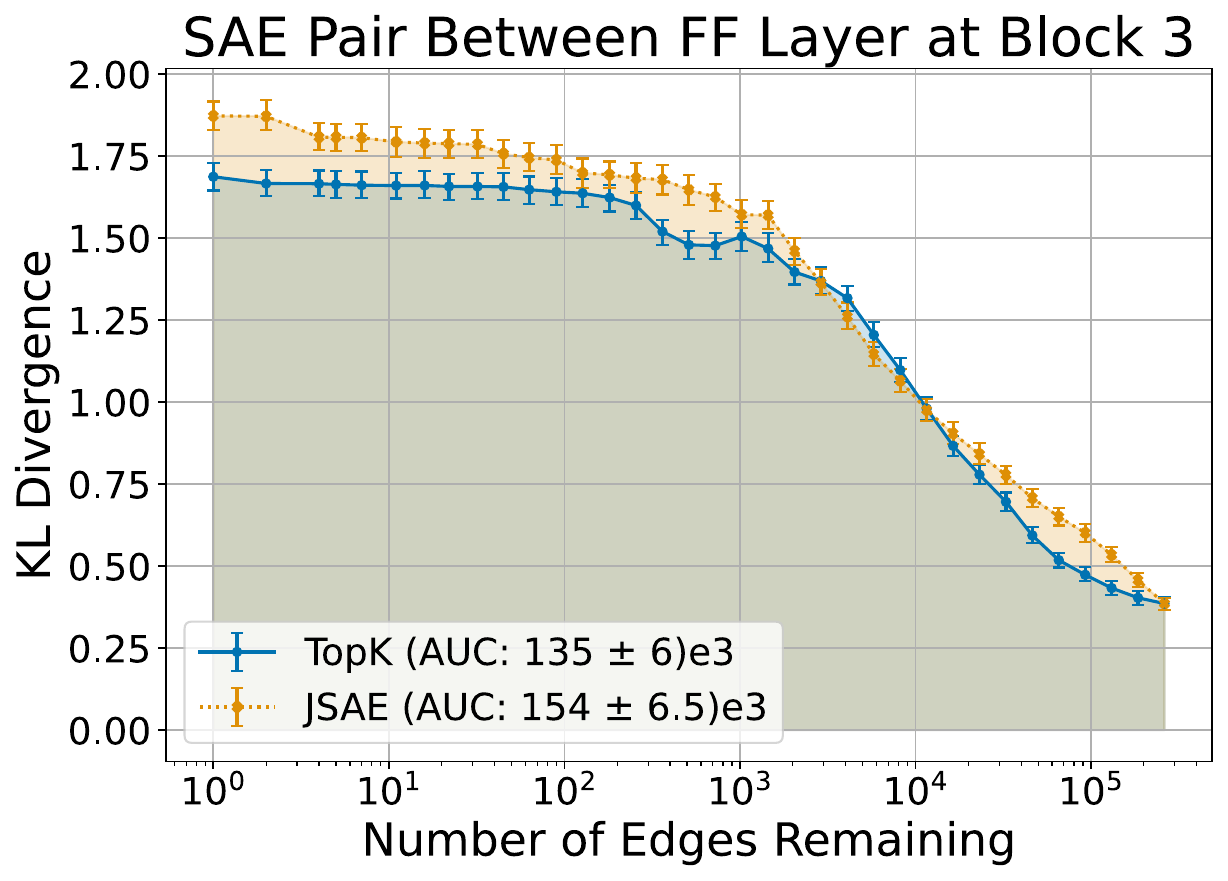}
    \end{subfigure}
    \hfill
    \begin{subfigure}[b]{0.31\textwidth}
        \centering
        \includegraphics[width=\textwidth]{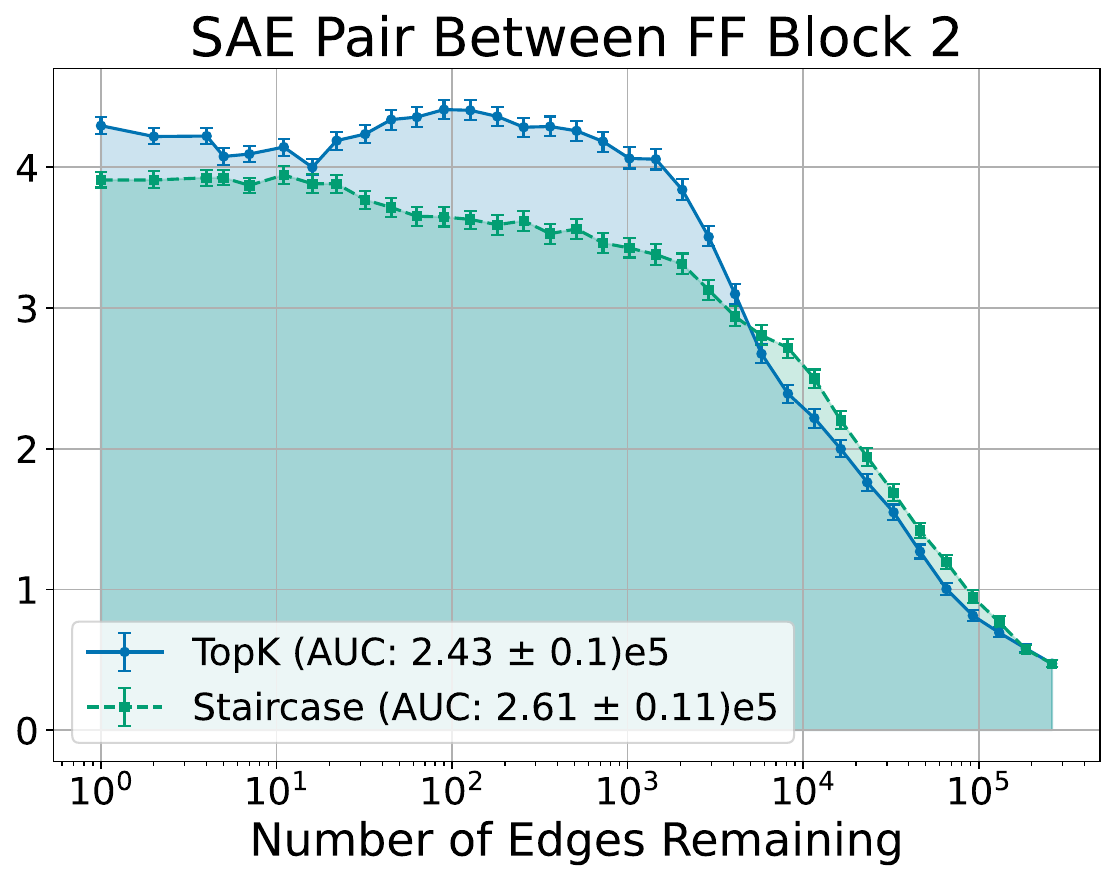}
    \end{subfigure}
    \hfill
    \begin{subfigure}[b]{0.31\textwidth}
        \centering
        \includegraphics[width=\textwidth]{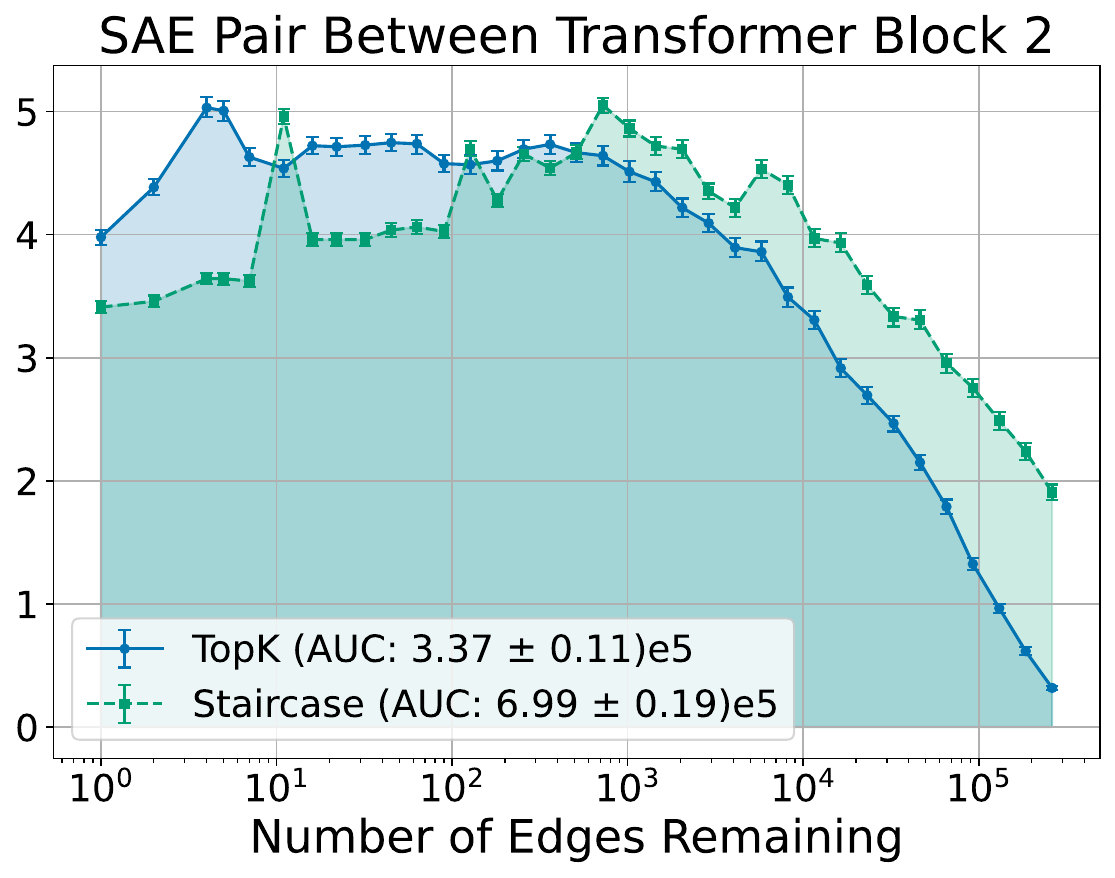}
    \end{subfigure}
    \caption{Ablation curves showing mixed performance cases where JSAE and Staircase SAEs do not clearly outperform standard TopK SAEs. These examples illustrate the variable performance across different compute blocks and highlight the importance of considering relative SCALAR scores that account for architectural differences.}
    \label{fig:three_figures_bad_appendix}
\end{figure}

While \Cref{fig:three_figures_good} in the main text shows representative cases where JSAE and Staircase SAEs clearly outperform TopK SAEs, performance varies across different compute blocks. \Cref{fig:three_figures_bad_appendix} shows examples where the methods do not clearly outperform standard TopK SAEs, demonstrating the variable nature of the improvements across different layers and architectural components.

\subsection{Complete Ablation Curves}

\begin{figure}[htbp]
	\centering
	\begin{subfigure}[b]{0.48\textwidth}
		\centering
		\includegraphics[width=\linewidth]{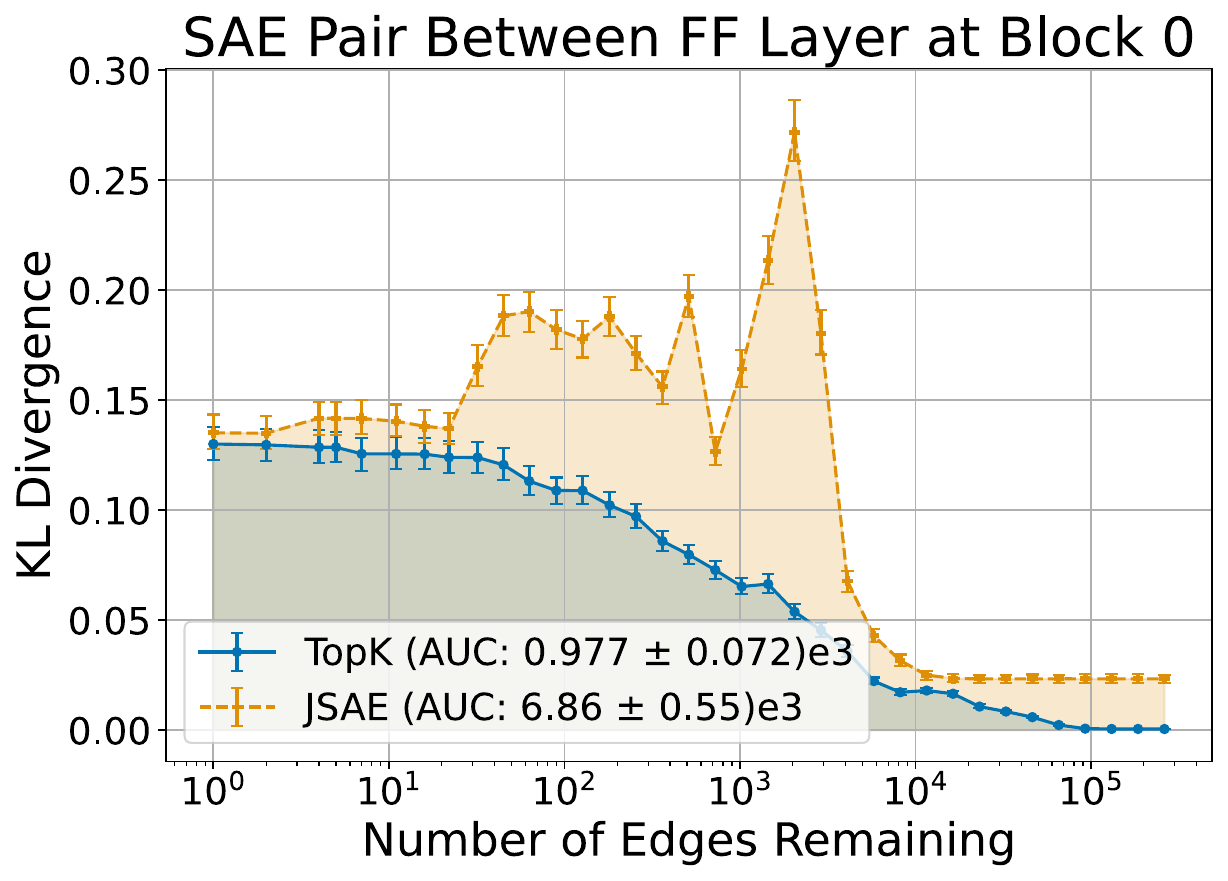}
		\label{fig:mlp_layer_0}
	\end{subfigure}
	\hspace{0.02\textwidth}
	\begin{subfigure}[b]{0.48\textwidth}
		\centering
		\includegraphics[width=\linewidth]{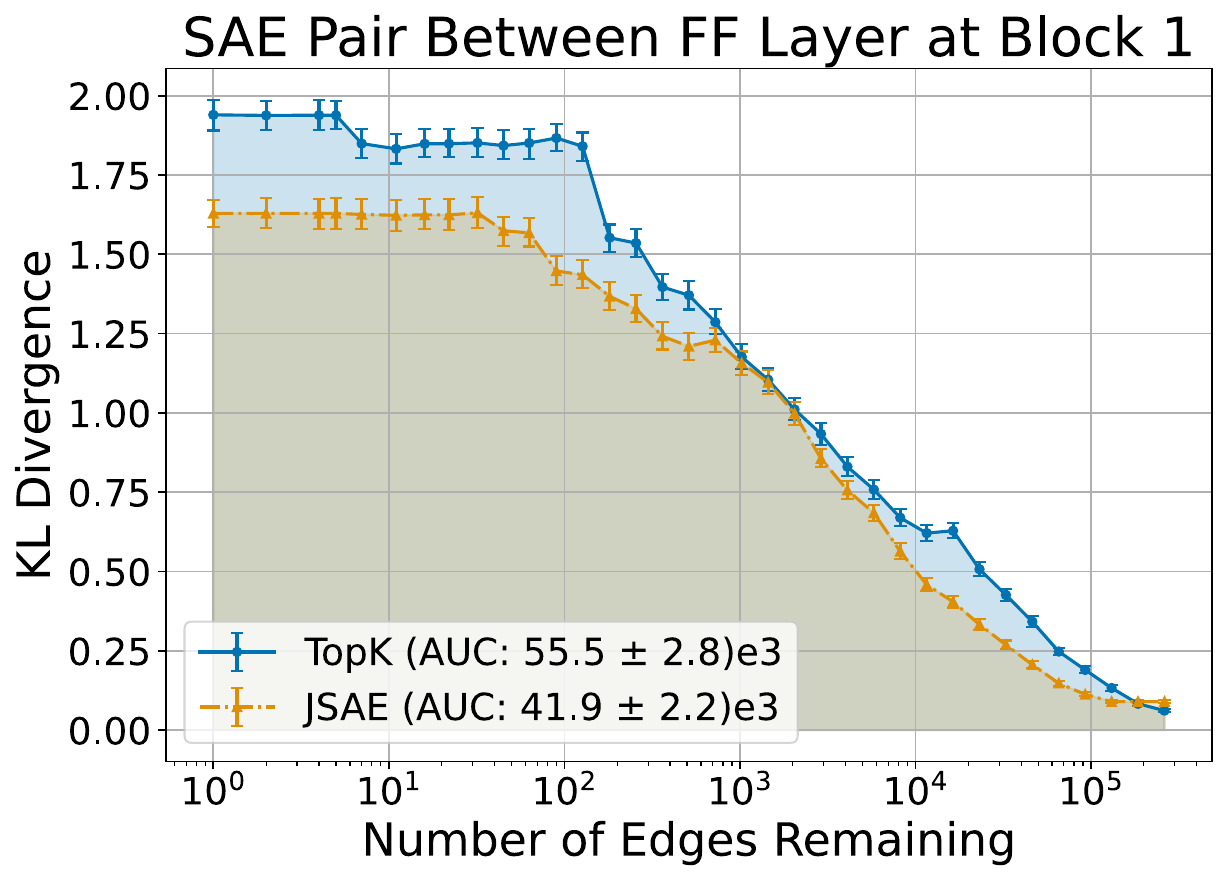}
		\label{fig:mlp_layer_1}
	\end{subfigure}
	\vskip\baselineskip
        \vspace{-0.5cm}
	\begin{subfigure}[b]{0.48\textwidth}
		\centering
		\includegraphics[width=\linewidth]{images/MLP_LAYER_2_updated.pdf}
		\label{fig:mlp_layer_2}
	\end{subfigure}
	\hspace{0.02\textwidth}
	\begin{subfigure}[b]{0.48\textwidth}
		\centering
		\includegraphics[width=\linewidth]{images/MLP_LAYER_3_updated.pdf}
		\label{fig:mlp_layer_3}
	\end{subfigure}
        \vspace{-0.35cm}
	\caption{Ablation curves for SAE pairs TopK \& JSAE between the feedforward (FF) network layer within each transformer block. The KL divergence is evaluated for edges logarithmically spaced from $1$ to $512 \times 512$, the total number of edges between each SAE pair. Besides the outlier at block zero, we observe similar performance on our metrics between TopK \& JSAE variants, indicating the tension between sparsity and reconstruction loss.}
	\label{fig:MLP_layer_ablation}
\end{figure}

\begin{figure}[htbp]
	\centering
	\begin{subfigure}[b]{0.48\textwidth}
		\centering
		\includegraphics[width=\linewidth]{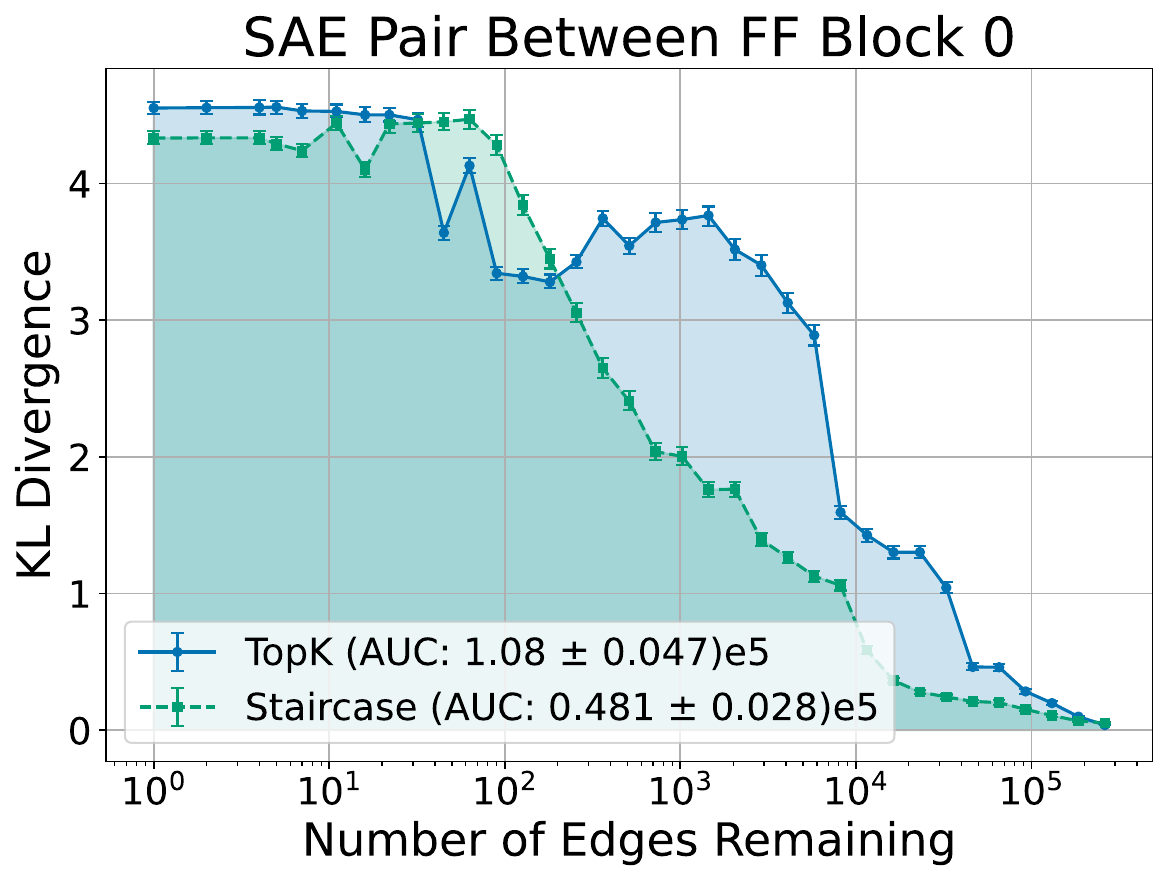}
		\label{fig:mlp_block_0}
	\end{subfigure}
	\hspace{0.02\textwidth}
	\begin{subfigure}[b]{0.48\textwidth}
		\centering
		\includegraphics[width=\linewidth]{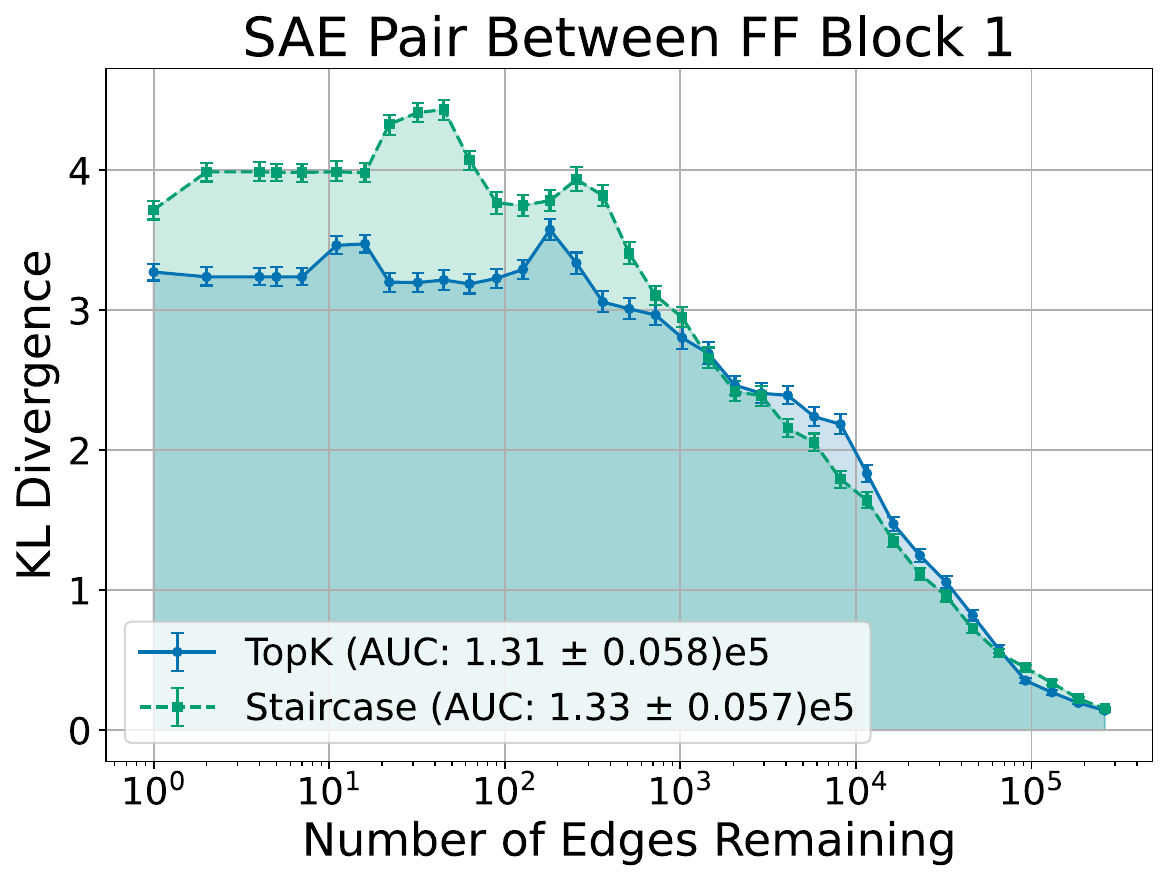}
		\label{fig:mlp_block_1}
	\end{subfigure}
	\vskip\baselineskip 
        \vspace{-0.5cm}
	\begin{subfigure}[b]{0.48\textwidth}
		\centering
		\includegraphics[width=\linewidth]{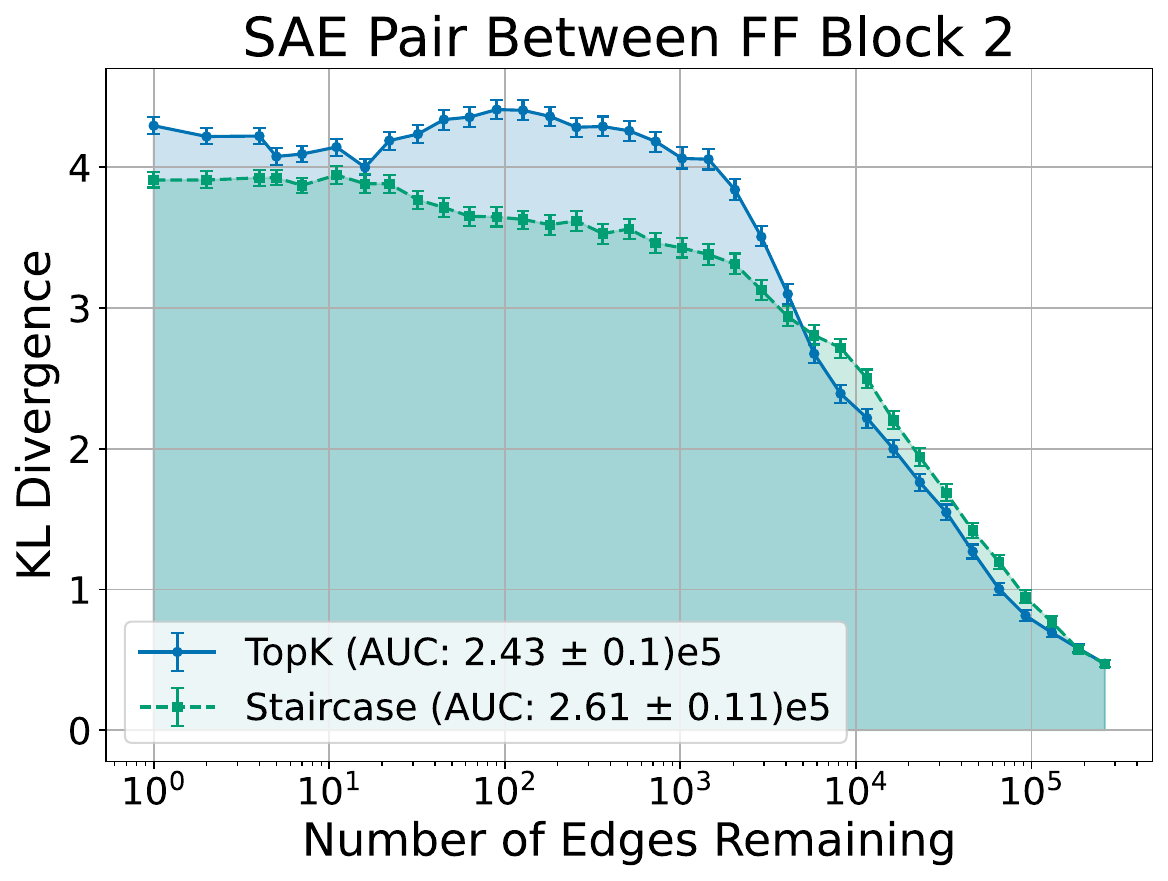}
		\label{fig:mlp_block_2}
	\end{subfigure}
	\hspace{0.02\textwidth}
	\begin{subfigure}[b]{0.48\textwidth}
		\centering
		\includegraphics[width=\linewidth]{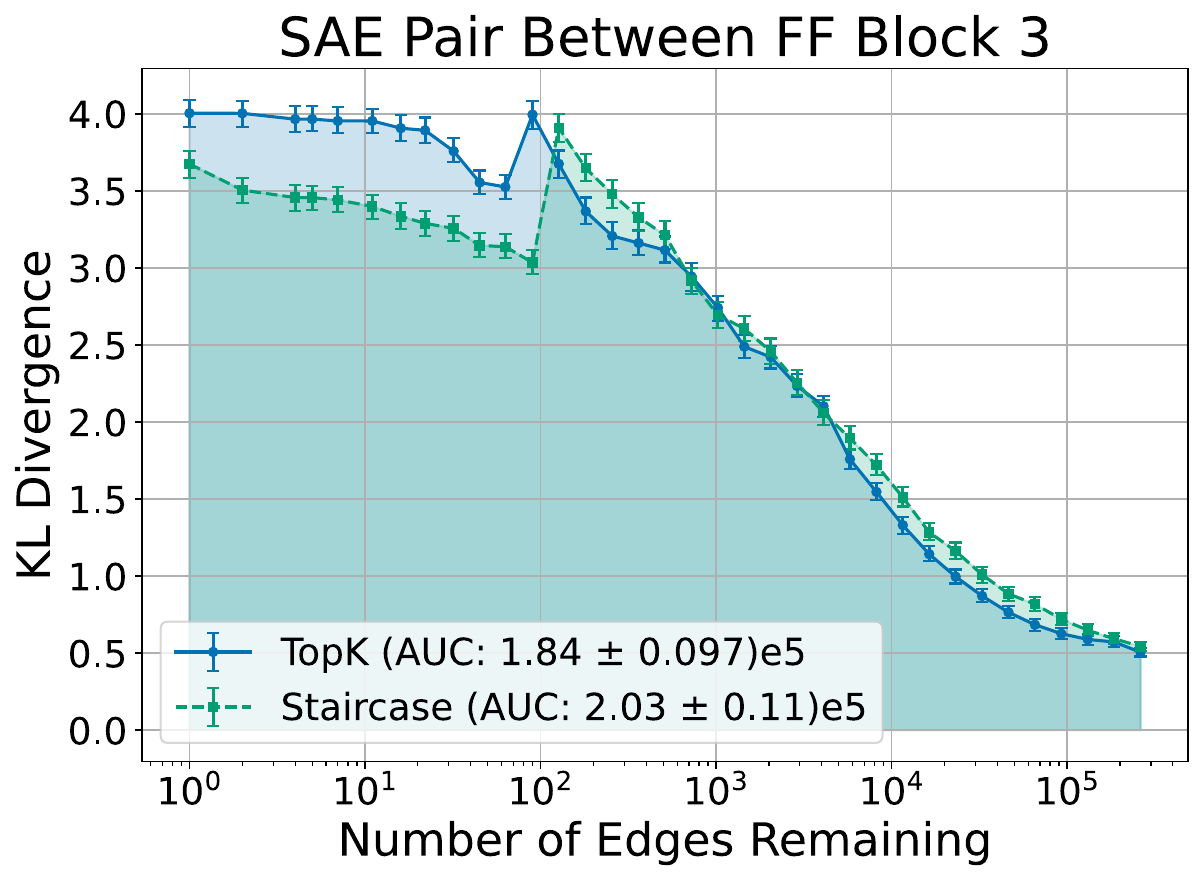}
		\label{fig:mlp_block_3}
	\end{subfigure}
        \vspace{-0.35cm}
	\caption{Ablation curves for SAE pairs TopK \& Staircase between the feedforward (FF) network block within each of the transformer blocks. The KL divergence is evaluated for edges logarithmically spaced from $1$ to $512 \times 512$, the total number of edges between each TopK SAE pair and half that Staircase pairs. Besides the outlier at block zero, we observe similar performance on absolute SCALAR score between TopK \& Staircase variants, while, Staircase outperforms TopK on relative SCALAR score across the board.}
	\label{fig:MLP_block_ablation}
\end{figure}

\begin{figure}[htbp]
	\centering
	\begin{subfigure}[b]{0.48\textwidth}
		\centering
		\includegraphics[width=\linewidth]{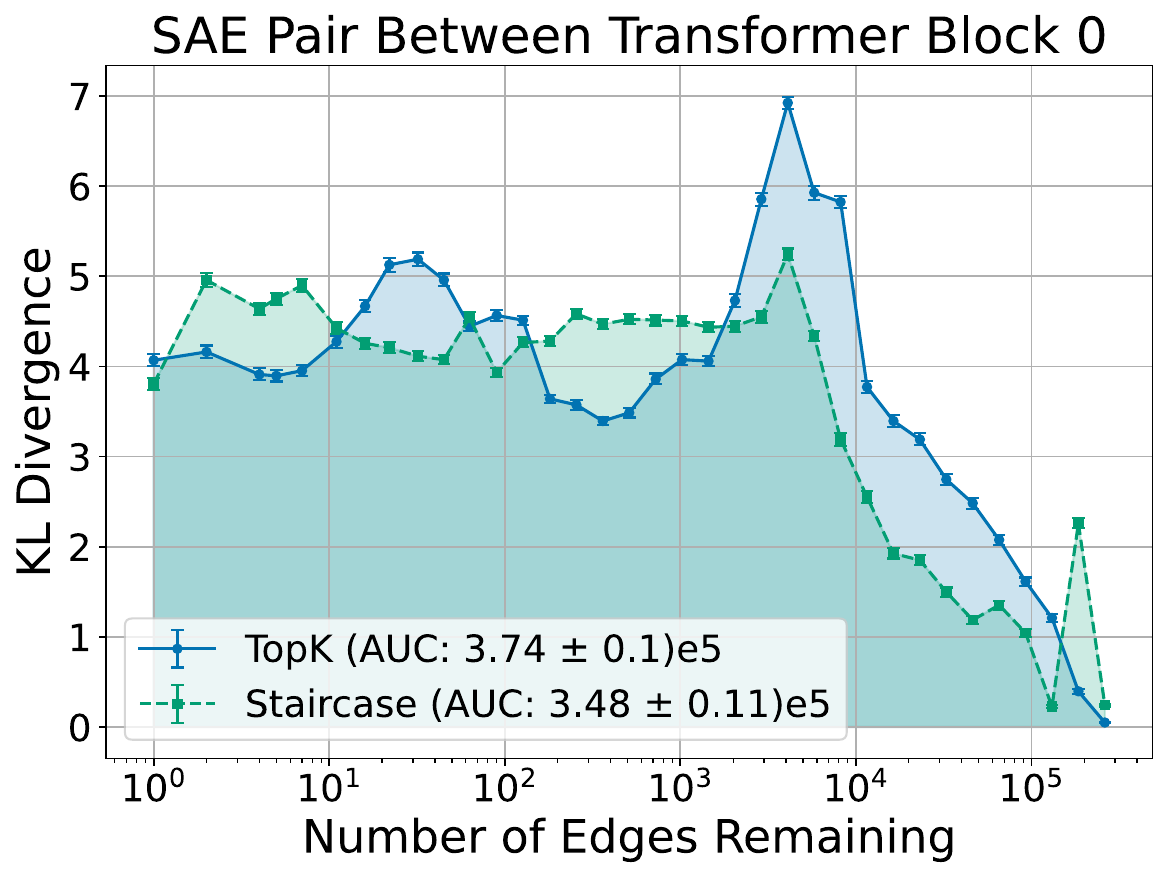}
		\label{fig:transformer_block_0}
	\end{subfigure}
	\hspace{0.02\textwidth}
	\begin{subfigure}[b]{0.48\textwidth}
		\centering
		\includegraphics[width=\linewidth]{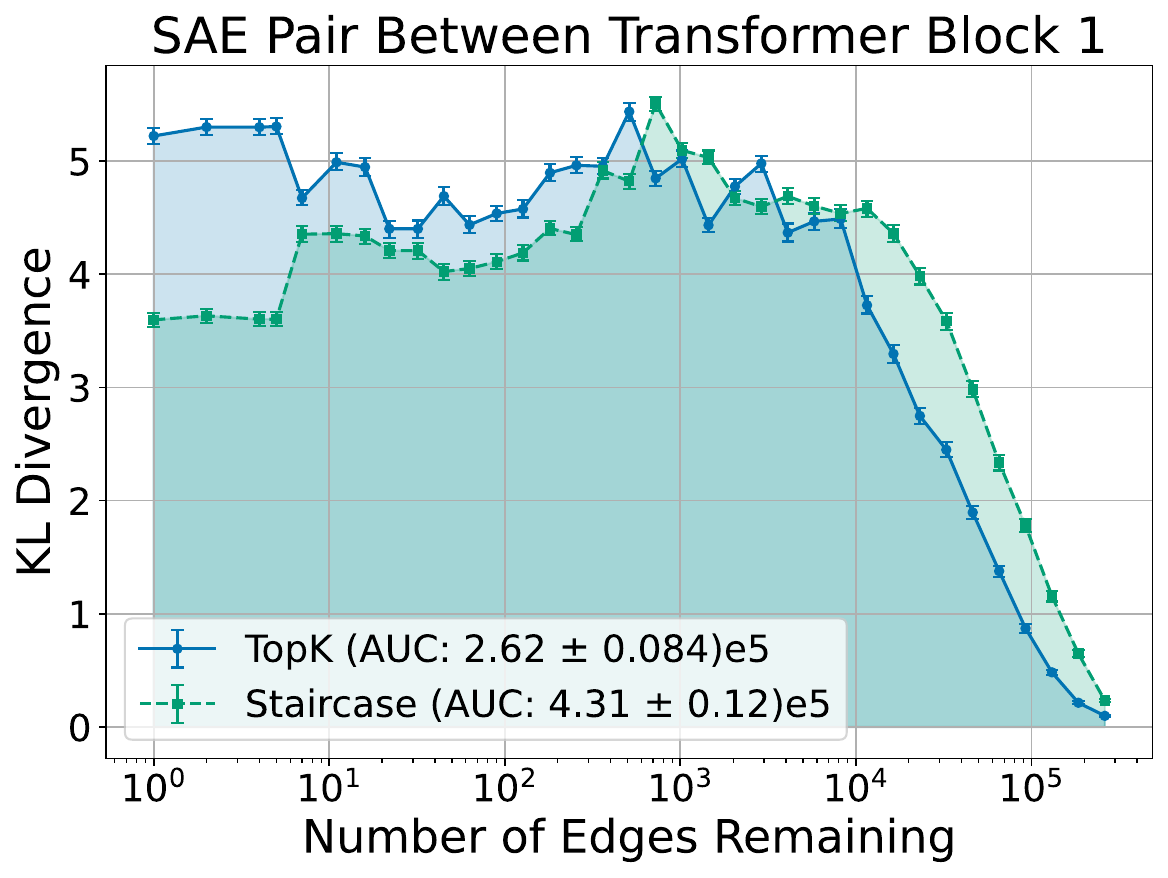}
		\label{fig:transformer_block_1}
	\end{subfigure}
	\vskip\baselineskip
        \vspace{-0.55cm}
	\begin{subfigure}[b]{0.48\textwidth}
		\centering
		\includegraphics[width=\linewidth]{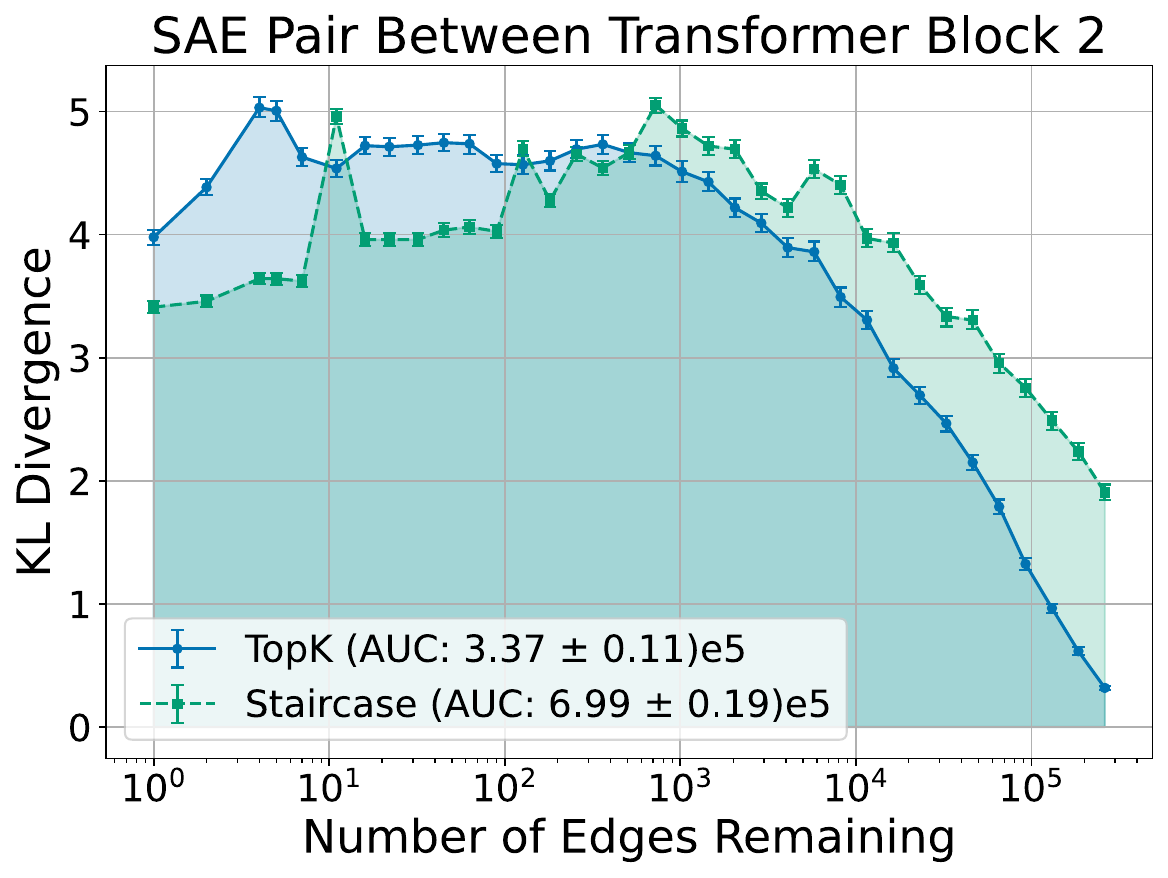}
		\label{fig:transformer_block_2}
	\end{subfigure}
	\hspace{0.02\textwidth}
	\begin{subfigure}[b]{0.48\textwidth}
		\centering
		\includegraphics[width=\linewidth]{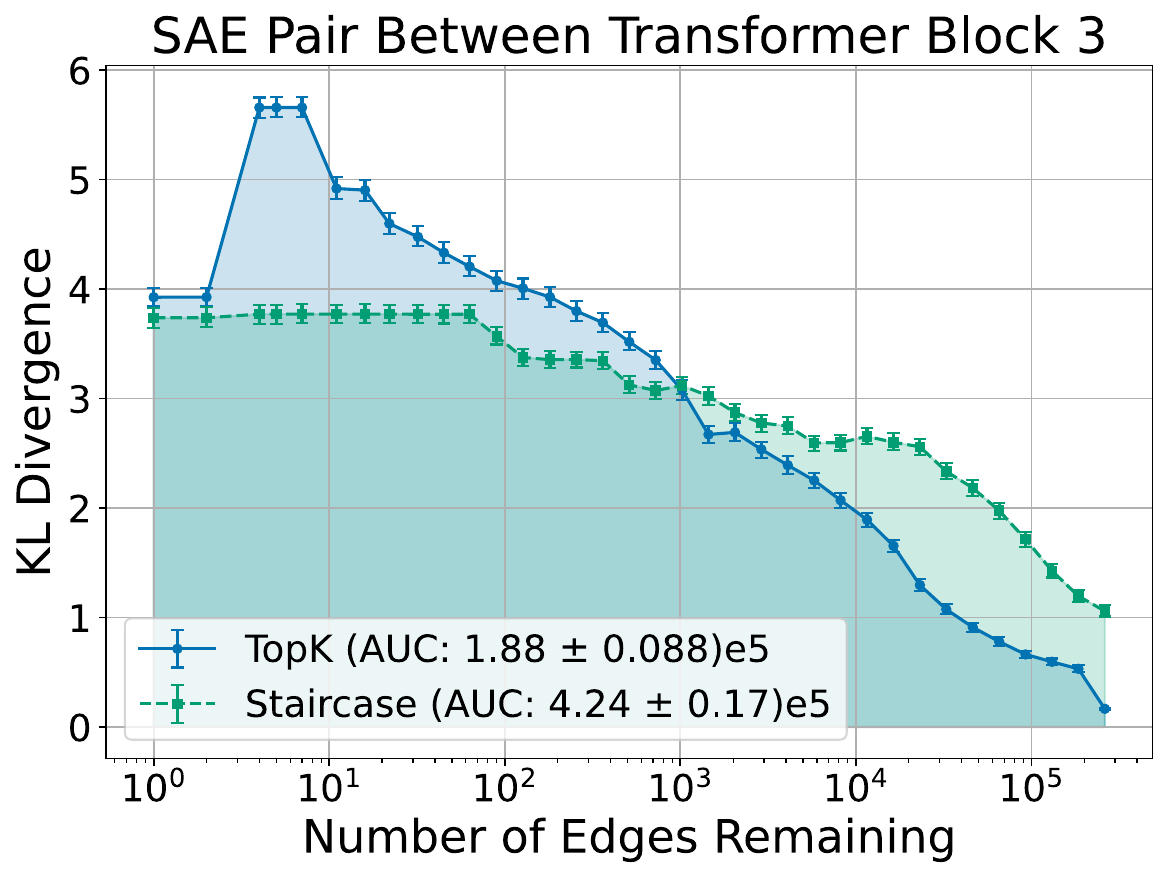}
		\label{fig:transformer_block_3}
	\end{subfigure}
        \vspace{-0.35cm}
	\caption{Ablation curves for SAE pairs TopK \& Staircase between each transformer  block. The KL divergence is evaluated for edges logarithmically spaced from $1$ to $512 \times 512$, the total number of edges between each TopK SAE pair. The number of edges between each Staircase pair at block $k$ is given by $512 \times 512 \times (k+1) \times (k+2)$. Besides the outlier at block zero, we observe TopK outperforms Staircase on absolute SCALAR score. While, at each position, Staircase outperforms TopK on relative SCALAR score.}
	\label{fig:Transformer_ablation}
\end{figure}

Fixing the edge number sequence:
\begin{gather}
    (1, 2, 4, 5, 7, 11, 16, 22, 32, 45, 63, 90, 127, 181, 256, 362, 512, 724, \nonumber\\
    1024, 1448, 2048, 2896, 4095, 5792, 8191, 11585, 16383, 23170, \label{equ:edge_count_sequence}\\
    32768, 46340, 65536, 92681, 131072, 185363, 262144) \nonumber .
\end{gather}
We produce ablation curves of each SAE variant for each block with a batch of 50 prompts across the locations; feedforward layer, feedforward block and transformer block,  see Figure \ref{fig:MLP_layer_ablation}, Figure \ref{fig:MLP_block_ablation} and Figure \ref{fig:Transformer_ablation} respectively.

\FloatBarrier
\subsection{Summary Tables}

In Table \ref{tab:abs_scalar_scores}, we collect the absolute SCALAR scores of each SAE variant across each block and for each SAE location.

\begin{table}[htbp]
    \centering
    \begin{tabular}{c||c|c|c|c||c}
         & Layer 0 ($10^{4}$) & Layer 1 ($10^{4}$) & Layer 2 ($10^{5}$) & Layer 3 ($10^{5}$) & \\
         \hline\hline
         TopK & $0.0977\pm 0.0072$ & $5.55\pm 0.28$ & $1.57\pm 0.06$ & $1.35\pm 0.06$ & FF Layer \\
         JSAE & $0.686\pm 0.055$ & $4.19\pm 0.22$ & $1.42\pm 0.06$ & $1.54\pm 0.07$ & \\ \hline
         TopK & $10.8\pm 0.5$ & $13.1\pm 0.6$ & $2.43\pm 0.10$ & $1.84\pm 0.10$ & FF Block\\
         Staircase & $4.81\pm 0.28$ & $13.3\pm 0.6$ & $2.61\pm 0.11$ & $2.03\pm 0.11$ & \\ \hline
         TopK & $37.4\pm 1.0$ & $26.2\pm 0.8$ & $3.37\pm 0.11$ & $1.88\pm 0.09$ & Trans. Block\\
         Staircase & $34.8\pm 1.1$ & $43.1\pm 1.2$ & $6.99\pm 0.19$ & $4.24\pm 0.17$ & \\
    \end{tabular}
    \caption{Absolute SCALAR scores derived from ablation plots.}
    \label{tab:abs_scalar_scores}
\end{table}

In Table \ref{tab:rel_scalar_scores}, we collect the relative SCALAR scores of each SAE variant across each block and for each location.

\begin{table}[htbp]
\centering
\begin{tabular}{c||c|c|c|c||c}
  & Layer 0 ($10^{4}$) & Layer 1 ($10^{4}$) & Layer 2 ($10^{5}$) & Layer 3 ($10^{5}$) & $(\times 2^{18})$ \\
\hline\hline
TopK & $0.0977\pm 0.0072$ & $5.55\pm 0.28$ & $1.57\pm 0.06$ & $1.35\pm 0.06$ & FF Layer \\
JSAE & $0.686\pm 0.055$ & $4.19\pm 0.22$ & $1.42\pm 0.06$ & $1.54\pm 0.07$ & \\ 
\hline
TopK & $10.8\pm 0.5$ & $13.1\pm 0.6$ & $2.43\pm 0.10$ & $1.84\pm 0.10$ & FF Block\\
Staircase & $2.40\pm 0.14$ & $6.65\pm 0.30$ & $1.30\pm 0.06$ & $1.01\pm 0.05$ & \\ 
\hline
TopK & $37.4\pm 1.0$ & $26.2\pm 0.8$ & $3.37\pm 0.11$ & $1.88\pm 0.09$ & Trans. Block\\
Staircase & $17.40\pm 0.55$ & $7.18\pm 0.20$ & $0.58\pm 0.02$ & $0.21\pm 0.01$ & \\ 
\end{tabular}
    \caption{Relative SCALAR scores derived from ablation plots.}
    \label{tab:rel_scalar_scores}
\end{table}

We emphasise that, given the chosen units each of Table \ref{tab:rel_scalar_scores}, the entries of the TopK and JSAE rows are the same as those appearing in Table \ref{tab:abs_scalar_scores}. In contrast, the Staircase rows in Table \ref{tab:abs_scalar_scores} are significantly smaller than those in Table \ref{tab:rel_scalar_scores}. This reflects that fact that  Staircase SAE pairs have more total edges than TopK and JSAE pairs.

\FloatBarrier
\section{GPT-2 Results}
\label{app:GPT-2}
Our results can be strengthened by demonstrating the applicability of both the SCALAR metric and the Staircase SAE performance on GPT-2 sized models. To this end, we have trained a suite of TopK and Staircase SAE pairs about the FF block of GPT-2 Small (124M parameters). To accommodate the additional scale, we have done the following.
\begin{enumerate}
\item Implemented a down-sampled version of integrated gradients. For this, we reduce the number of intervals used to approximate the integrals for gradient attribution and the number of activations used in total.
\item Implemented a down-sampled ablation study on an early layer (layer 1), a middle layer (layer 6) and a later layer (layer 11). For these, we perform ablations for each edge number in~$\{10^{0}, 10^{2},...,10^{6}\}$, on three distinct prompts.
\end{enumerate}
For this modified setup, we present a summary of results in Table~\ref{tab:scalar_feedforwardblock_gpt2}. We find that the Staircase SAEs provide a $38.69 \pm 0.71 \%$ improvement over TopK in relative interaction sparsity (relative SCALAR score), see Section \ref{subsec:STgpt2} for the full set of results. Contrasting this result to the analogous $59.67 \pm 1.83 \% $ reduction in the toy setup, we believe that this provides preliminary evidence that our results can be applied to realistic model architectures. 

\begin{table}[h]
    \centering
    \resizebox{\textwidth}{!}{\begin{tabular}{lcccc}
        \toprule
        & \textbf{Block 1} & \textbf{Block 6} & \textbf{Block 11} & \textbf{Aggregate} \\
        \midrule
        \textbf{Absolute reduction (\%)} & $ -5.71\pm 2.25$ & $ -8.73\pm 2.33$ & $ -55.59\pm 2.85$ & $ -22.62\pm 1.42$ \\
        \textbf{Relative reduction (\%)} & $ 47.15\pm 1.13$ & $ 45.64\pm 1.16$ & $ 22.21\pm 1.43$ & $ 38.69\pm 0.71$ \\
        \bottomrule
    \end{tabular}}
    \vspace{0.5em}
	\caption{Percentage reduction in SCALAR scores for Staircase SAEs compared to TopK SAEs across feedforward blocks of GPT-2 Small. Positive values indicate improved sparsity (lower SCALAR score) for Staircase SAEs.}
        \label{tab:scalar_feedforwardblock_gpt2}
\end{table}

\begin{figure}[t!]
    \centering
    \begin{subfigure}[b]{0.32\textwidth}
        \centering
        \includegraphics[width=\textwidth]{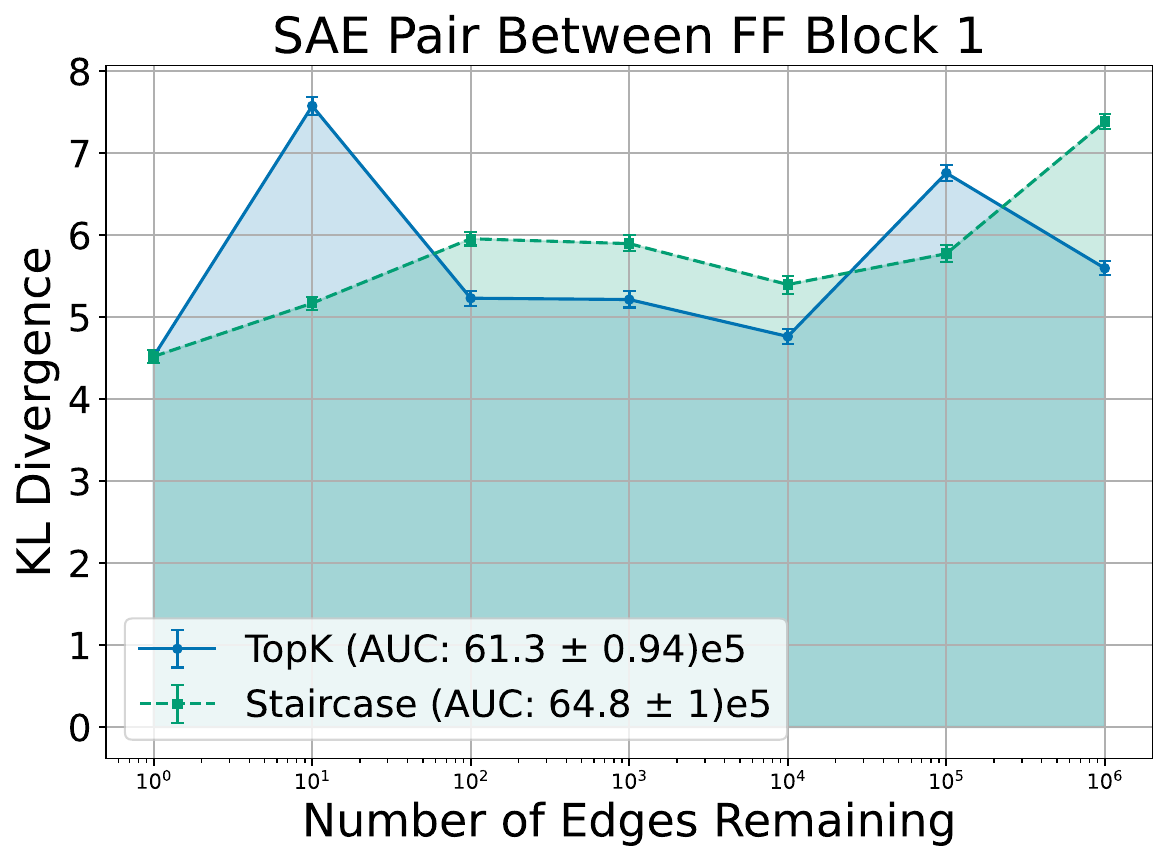}
    \end{subfigure}
    \hfill
    \begin{subfigure}[b]{0.32\textwidth}
        \centering
        \includegraphics[width=\textwidth]{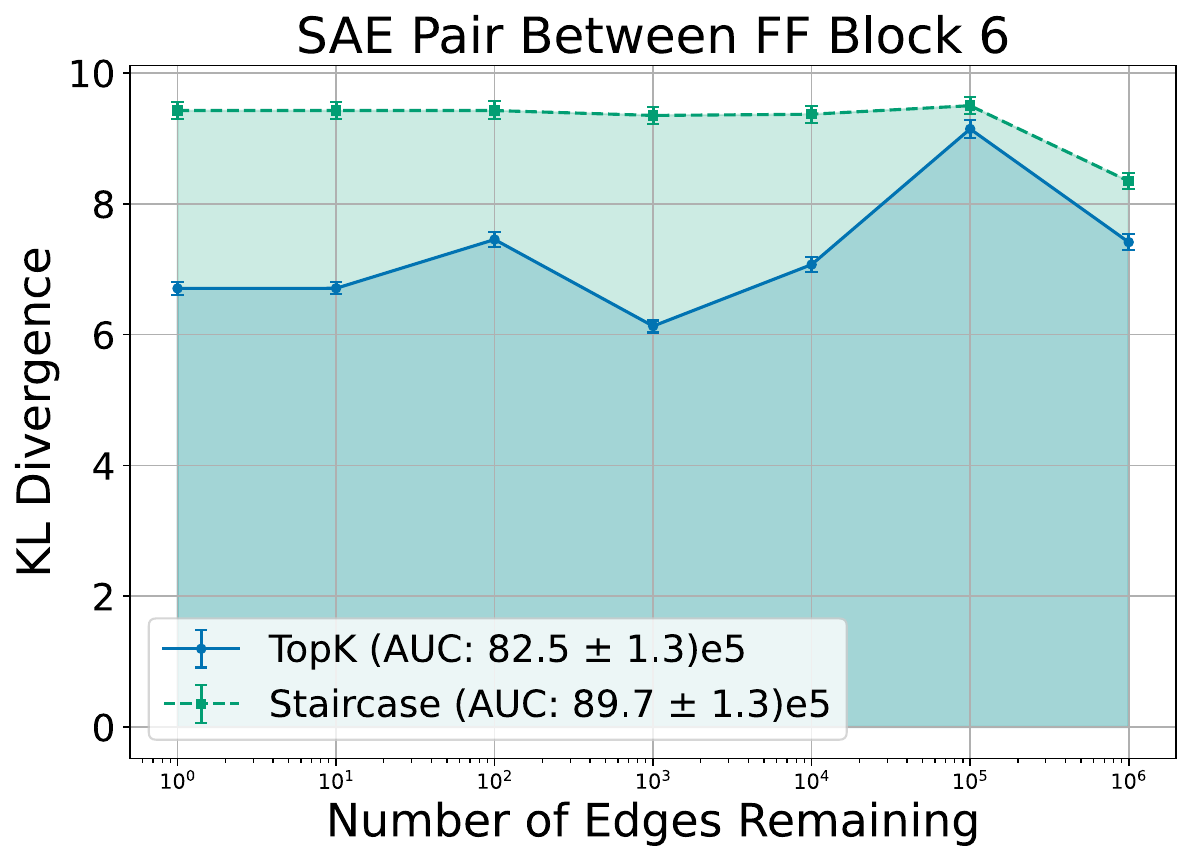}
    \end{subfigure}
    \hfill
    \begin{subfigure}[b]{0.32\textwidth}
        \centering
        \includegraphics[width=\textwidth]{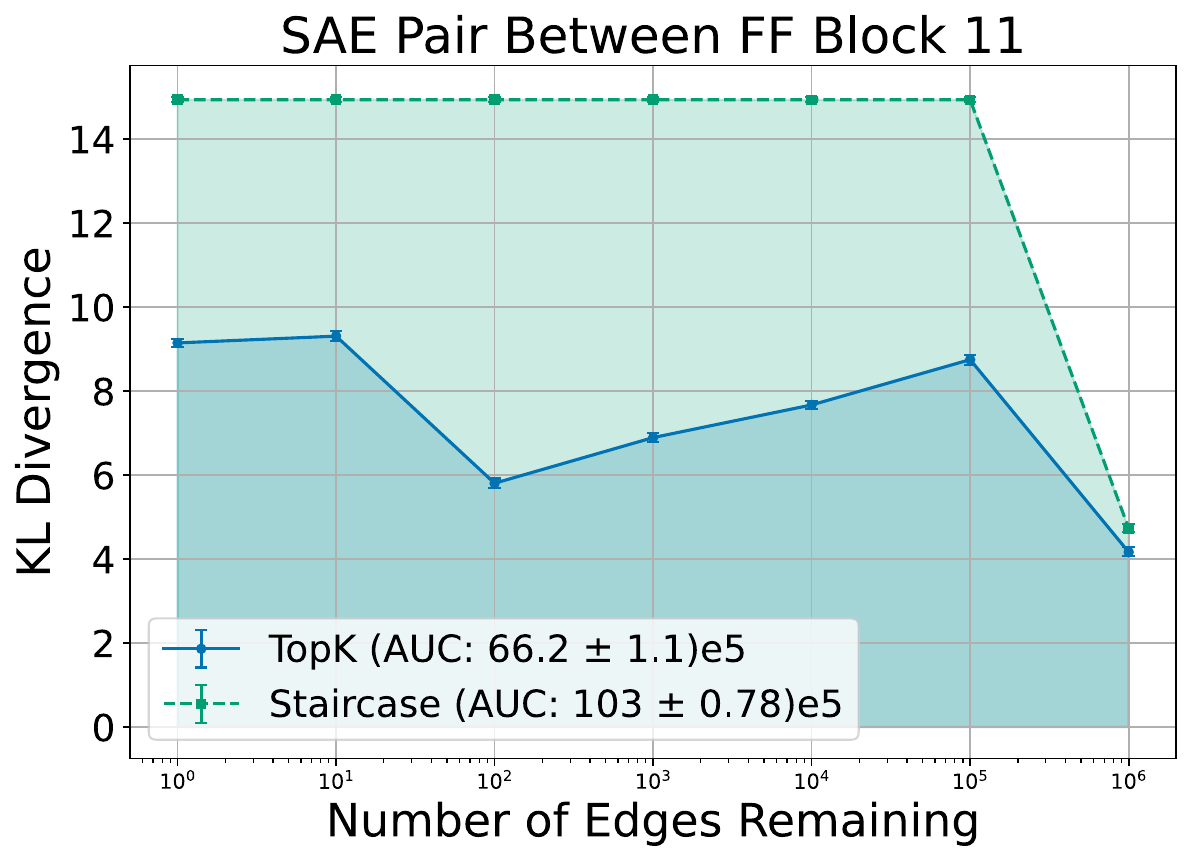}
    \end{subfigure}
    \caption{The ablation curves for SAEs attached at the feedforward block in the GPT-2 Small model.}
    \label{fig:gpt-2}
\end{figure}

As expected, the absolute SCALAR scores (Table~\ref{tab:abs_scalar_scores_gpt2}) are higher for both architectures at this scale, with Staircase SAEs showing higher raw AUC values due to their increased connectivity, see Figure \ref{fig:gpt-2} for the ablation plots. The reported improvements are measured using relative SCALAR scores (Table~\ref{tab:rel_scalar_scores_gpt2}), which normalize by the number of potential connections to enable fair architectural comparison.

\subsection{Summary tables}
\label{subsec:STgpt2}

In Table \ref{tab:abs_scalar_scores_gpt2}, we collect the absolute SCALAR scores of each SAE variant across the feedforward blocks 1, 6 and 11.

\begin{table}[htbp]
    \centering
    \begin{tabular}{c||c|c|c}
         & Layer 1 ($10^{5}$) & Layer 6 ($10^{5}$) & Layer 11 ($10^{5}$) \\
         \hline\hline
         TopK & $ 61.3\pm 0.9$ & $ 82.5\pm 1.3$ & $66.2 \pm 1.1$ \\
         Staircase & $ 64.8\pm 1.0$ & $89.7 \pm 1.3$ & $ 103.0\pm 0.8$
    \end{tabular}
    \caption{Absolute SCALAR scores derived from GPT-2 Small ablation plots of the feedforward block.}
    \label{tab:abs_scalar_scores_gpt2}
\end{table}

In Table \ref{tab:rel_scalar_scores_gpt2}, we collect the relative SCALAR scores of each SAE variant across the feedforward blocks 1, 6 and 11.
\begin{table}[htbp]
    \centering
    \begin{tabular}{c||c|c|c}
         ($ 2^{26} \times 3^{2} \times $) & Layer 1 ($10^{5}$) & Layer 6 ($10^{5}$) & Layer 11 ($10^{5}$) \\
         \hline\hline
         TopK & $ 61.3\pm 0.9$ & $ 82.5\pm 1.3$ & $66.2 \pm 1.1$\\
         Staircase & $ 32.4\pm 0.5$ & $44.9 \pm 0.7$ & $ 51.5\pm 0.4$
    \end{tabular}
    \caption{Relative SCALAR scores derived from GPT-2 Small ablation plots of the feedforward block.}
    \label{tab:rel_scalar_scores_gpt2}
\end{table}

\FloatBarrier
\section{Impact Statement}
\label{app:ImpactStatement}

In this work, we contribute to the mechanistic interpretability literature by developing techniques to identify and promote computational sparsity among SAE variants. To this end, we introduce a benchmark and a novel SAE architecture. We envisage practitioners using our benchmark to develop computationally sparse SAE architectures. We also anticipate circuit analysis to be more tractable on our SAE variant compared to alternatives. By promoting computational sparsity, we strive towards a more complete understanding of language models, mitigating concerns such as algorithmic bias and manipulation.

\end{document}